%% file: main.tex
\documentclass[runningheads]{llncs}
\usepackage[T1]{fontenc}
\usepackage{graphicx}
\usepackage{booktabs}
\usepackage[misc]{ifsym}

\usepackage{algorithm}
\usepackage{algorithmic}

\usepackage{latexsym}
\usepackage{amssymb}
\usepackage{amsmath}
\usepackage{soul}
\usepackage{enumitem}
\usepackage{booktabs}
\usepackage{makecell}
\allowdisplaybreaks

\usepackage{color} 
\usepackage{subcaption}
\usepackage{tikz}
\usepackage{pgfplots}
\pgfplotsset{compat=1.18}
\usepgfplotslibrary{groupplots,fillbetween}
\usetikzlibrary{patterns}        
\usetikzlibrary{calc}

\pgfplotsset{
  overview/.style={
    width=0.40\linewidth,
    height=0.22\linewidth,
    grid=both,
    grid style={line width=.1pt},
    major grid style={line width=.2pt},
    tick label style={font=\scriptsize},
    label style={font=\scriptsize},
    title style={font=\scriptsize},
  },
  methodline/.style={thick},
  ablationmark/.style={only marks, mark=*, mark size=1.7pt},
}

\definecolor{darkgreen}{rgb}{0.0, 0.5, 0.0} 
\definecolor{darkred}{rgb}{0.65, 0.0, 0.0} 
\definecolor{darkblue}{rgb}{0.0, 0.0, 0.65} 
\definecolor{darkyellow}{rgb}{0.75, 0.6, 0.0} 
\definecolor{custompurple}{RGB}{148, 103, 189} 
\definecolor{darkorange}{RGB}{180, 120, 0} 
\definecolor{lightgreen}{rgb}{0.6, 1.0, 0.6}  
\definecolor{lightyellow}{rgb}{1.0, 1.0, 0.6}  

\newcommand{\refSuppMDP}{Appendix~A}          
\newcommand{\refSuppFeas}{Appendix~B}         
\newcommand{\refSuppGen}{Appendix~C}          
\newcommand{\refSuppSP}{Appendix~D}           
\newcommand{\refSuppDRL}{Appendix~E}          

\begin{document}

\title{Navigating Demand Uncertainty in Container Shipping: Deep Reinforcement Learning for Enabling Adaptive and Feasible Master Stowage Planning}

\titlerunning{Navigating Demand Uncertainty in Container Shipping}

\author{
    Jaike van Twiller\inst{1} (\Letter), 
    Yossiri Adulyasak\inst{2,3}, 
    Erick Delage\inst{2,3,4}, \\
    Djordje Grbic\inst{1}, 
    Rune Møller Jensen\inst{1}
}
\institute{
    IT University of Copenhagen, Denmark 
    \\ \email{jaiv@itu.dk}
    \and
    HEC Montréal, Qc, Canada
    \and
    GERAD, Qc, Canada
    \and
    Mila - Quebec AI Institute, Qc, Canada
}
\authorrunning{Van Twiller et al.}




\toctitle{Navigating Demand Uncertainty in Container Shipping: Deep Reinforcement Learning for Enabling Adaptive and Feasible Master Stowage
Planning}
\tocauthor{Jaike van Twiller, Yossiri Adulyasak, Erick Delage, Djordje Grbic, Rune Møller Jensen}

\maketitle              

\begin{abstract}
Reinforcement learning (RL) has successfully solved various deterministic and stochastic planning problems. However, conventional RL struggles with complex real-world constraints, particularly when feasibility is explicit and depends on the current state or trajectory. In this work, we address stochastic sequential decision-making with state-dependent constraints through a real-world case study of the master stowage planning problem in container shipping, which aims to optimize revenue and costs under demand uncertainty and operational constraints. We propose a deep RL framework with an encoder-decoder model that integrates problem instance, solution, and uncertainty information to guide planning. We introduce differentiable projection layers that enforce convex polyhedral constraints, while Jacobian corrections offset the projections to yield unbiased policy gradient estimates. Experiments show that our model efficiently finds adaptive, feasible solutions that generalize across distribution shifts and scale to longer planning horizons, outperforming state-of-the-art baselines in constrained RL and stochastic programming. As such, our policies enable adaptive, uncertainty-aware planning that can support resilient and sustainable supply chains.
\end{abstract}

\section{Introduction}\label{sec:intro}
In recent years, reinforcement learning (RL) has performed well on planning benchmarks such as vehicle routing and scheduling \cite{bengioMachineLearningCombinatorial2021,mazyavkinaReinforcementLearningCombinatorial2021,koolAttentionLearnSolve2019,hottungEfficientActiveSearch2022,kwonPOMOPolicyOptimization2020}. Yet these benchmarks are often deterministic with implicit feasibility, whereas real-world decision problems are stochastic, sequential, and deal with state/trajectory-dependent hard constraints. In this setting, traditional stochastic programs based on explicit scenario trees can become computationally expensive \cite{powellReinforcementLearningStochastic2022}, while many constrained-RL methods enforce approximate feasibility \cite{zhangPenalizedProximalPolicy2022}. Instead, exact per-step projection onto state-dependent polyhedra can be used but often exceeds operational time budgets \cite{agrawalDifferentiableConvexOptimization2019,Kasaura2023Benchmarking}, motivating fast feasibility mechanisms for longer-horizon planning under uncertainty.

We study this challenge in container shipping, which transports 45\% of global goods (\$8.1 trillion/year) \cite{unitednationsconferenceontradeanddevelopmentReviewMaritimeTransport2021} and emits $>$200 million tons of CO$_2$ per year \cite{lloydslistShippingEmissionsRise2022}, yet is central to decarbonization due to low emissions per ton-mile \cite{europeancommissionReducingEmissionsShipping2023}. Container shipping involves several constrained sequential decision problems with inherent uncertainty, e.g., berth allocation \cite{martin-iradiMultiportBerthAllocation2022}, pre-marshalling \cite{hottungDeepLearningAssisted2020}, quay crane scheduling \cite{herupLinearTimeAlgorithm2022}, and container vessel stowage planning \cite{vantwillerLiteratureSurveyContainer2024}. The latter can decompose into two challenging subproblems: the upstream master planning problem (MPP) assigns cargo to clusters of slots, and the downstream slot planning problem (SPP) allocates containers to individual slots \cite{pacinoFastGenerationNearoptimal2011}. As uncertainty drives frequent demand updates, fast re-planning is needed to assess the downstream feasibility of MPP decisions and support resilient, sustainable supply chains.

We propose a deep reinforcement learning (DRL) framework to generate feasible, high-profit solutions for the MPP under demand uncertainty. We formulate the problem as an action-constrained MDP and learn a planning policy that conditions on instance features, the current solution state, and demand realizations. The policy uses a projected encoder–decoder model, with feasibility enforced by fast differentiable projections onto state-dependent convex polyhedral constraint sets. Because actions are projected during training, we apply a Jacobian-based log-probability correction to obtain unbiased policy-gradient estimates. 

Our main contributions are:
\begin{itemize}
    \item \textbf{MPP environment:} A novel MDP formulation for MPP under demand uncertainty with state-dependent coupled constraints, integrating stochastic revenue management and stowage decisions. The benchmark is released as open-source to support reproducible evaluation and mitigate data scarcity.\footnote{\url{https://github.com/OptimalPursuit/navigating\_uncertainty\_in\_mpp}}
    \item \textbf{Fast feasibility projection:} We propose a differentiable projection layer that minimizes violations of state-dependent coupled convex polyhedral constraints, enabling feasible policy learning with tight computational budgets.
    \item \textbf{Adaptive planning under uncertainty:} Our model outperforms constrained-RL baselines and outpaces stochastic programs by efficiently generating feasible, high-profit plans that show robustness under distribution shift and scale to longer voyages. Ablation studies isolate the impact of key components.
\end{itemize}

\section{Domain Preliminaries}  \label{sec:domain}
We present some domain preliminaries to provide context. For a comprehensive review, we refer to \cite{jensenContainerVesselStowage2018}.

Liner shipping companies operate vessels on fixed-schedule, closed-loop voyages, much like a maritime bus service. These journeys, often exceeding ten ports, typically originate in supply-surplus regions (e.g., Asia) and finish in high-demand areas (e.g., Europe). A voyage can be described as a directed path graph $G_P = (P, E_P)$, where $P = \{1,2,\dots, N_P\}$ are the nodes representing ports, and $E_P  = \{(i,j) \in {P}^2\!\mid\! j = i + 1\}$ are the edges representing legs between ports. We define sub-voyages by the set ${P}_\text{start}^\text{end} = \{p \in {P}\!\mid\! \text{start}\leq p \leq \text{end}\}$. Containers have a port of load (\textit{pol}) and a port of discharge (\textit{pod}), named transports or origin-destination pairs. A transport $\textit{tr} = (\textit{pol},\textit{pod})$ is defined with $\textit{pol} \in {P}_1^{N_P\text{-}1}$ and $\textit{pod} \in {P}_{2}^{N_P}$. The set of all possible transports is $\textit{TR}=\{(i,j) \in {P}^2\!\mid\! i<j\}$.

A standard container unit is the twenty-foot equivalent unit (TEU). Container cargo can be categorized as follows: lengths (20 ft or 1 TEU, 40 ft or 2 TEU), weight classes (light, medium, heavy), and customer contracts (long-term, spot market) to determine cargo revenue ($\textit{rev}$). Long-term contracts offer lower but stable revenue and high volumes, taking priority over spot market contracts, which provide higher yet volatile revenue in lower volumes. The lack of no-show fees creates uncertainty in container demand, although this becomes more predictable closer to the arrival date. This necessitates dynamic cargo allocation to vessels during the voyage. Containers are often grouped into cargo classes by characteristics, such as ${K} = \{20\textit{ft}, 40\textit{ft}\} \times \{\textit{Light}, \textit{Medium}, \textit{Heavy}\} \times \{\textit{Spot},\textit{Long}\}$.

\begin{figure}[t!]
\centering
\includegraphics[scale=0.28]{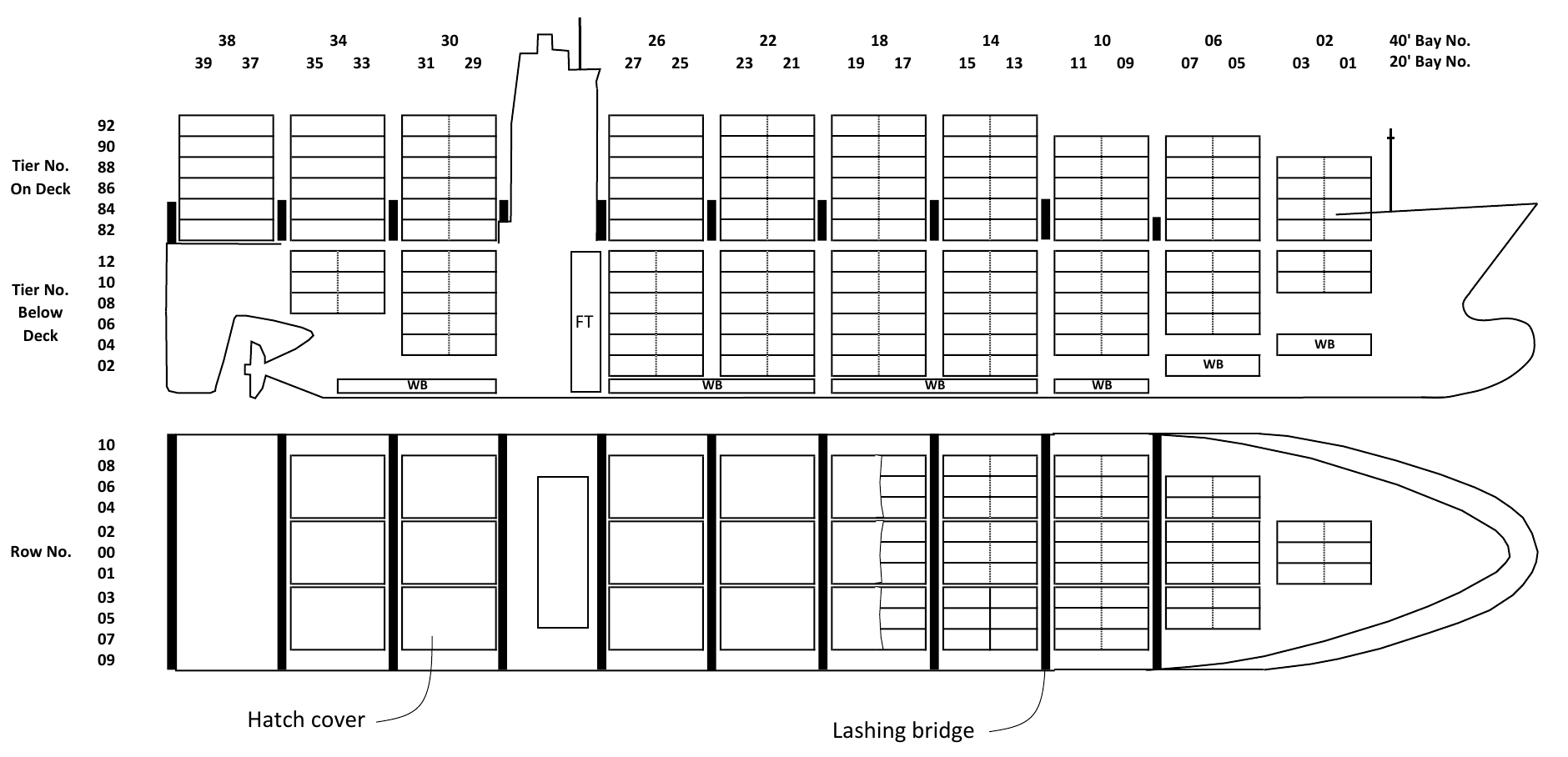}
\caption{The side and top view of a container vessel.}
\label{fig:vessel1}
\end{figure}

Figure \ref{fig:vessel1} shows the cellular layout of vessels with standard coordinates of \textit{(bay, row, tier)}. It is divided into bays (02-38) ranging from the front (e.g., bay 02) to the rear (e.g., bay 38), also named from fore to aft. Bays are defined by an ordered set ${B} = \{1,2,\dots,N_B\}$, ranging from fore to aft with $N_B$ being the last bay. Each bay contains stacks of cells, whereas cells contain two slots that can hold two 20-foot containers or one 40-foot container. Bays are horizontally separated by hatch covers into above and below deck sections, introducing the set of decks ${D} = \{d^\textit{above}, d^\textit{below}\}$. Note that hatch covers introduce complexity, as below-deck cargo can only be discharged if the above-deck cargo is cleared. 

Vessel utilization  \( u_p \in \mathbb{Z}^{|B| \times |D| \times |K| \times |\textit{TR}|}_{\geq 0} \) represents the number of containers at departure from port $p$ across bays $B$, decks $D$, cargo types $K$, and transports $\textit{TR}$. The number of containers loaded at port $p$ are defined by $u^+_p \in \mathbb{Z}_{\geq 0}^{|B| \times |D| \times |K| \times |\textit{TR}|}$, whereas the number of containers discharged at port $p$ are denoted by $u^-_p \in \mathbb{Z}_{\geq 0}^{|B| \times |D| \times |K| \times |\textit{TR}|}$. The utilization at port $p$ is defined as $u_p = u_{p-1} + u^+_p - u^-_p \; \forall p \in {P}$, where $u_0$ is the vessel's arrival condition at the first port. We also define the vessel's pre-loading utilization $u'_p = u_{p-1} - u^-_p$.

The MPP creates a capacity management plan for each general location (bay, deck) to maximize revenue and minimize costs, while the subsequent SPP assigns containers to specific slots. In multi-port voyages, each decision affects future feasibility and optimality. Due to frequent re-planning, runtimes over 10 minutes are found impractical \cite{pacinoFastGenerationNearoptimal2011}. Key efficiency objectives include minimizing overstowage (an NP-hard problem \cite{tierneyComplexityContainerStowage2014}) and crane makespan. Overstowage occurs when containers block access to those below that need to discharge, resulting in unnecessary moves. Minimizing makespan reduces port stay duration and associated costs. Furthermore, vessels must satisfy safety constraints for seaworthiness, particularly their weight distribution. The longitudinal (lcg) and vertical center of gravity (vcg) must remain within bounds for utilization $u_p$, where $\otimes$ denotes the outer product and $\odot$ the element-wise Hadamard product. All operands are vectorized before the product ($w$ is broadcast to match $u_p$), so $\mathbf{1}^\top(\cdot)$ sums over all entries. Hence, we get the following bounds for lcg and vcg:
\begin{align}
\underline{\textit{lcg}} \leq \frac{\mathbf{1}^\top \big( \textit{lm} \odot u_p \big)}{\mathbf{1}^\top \big( w \odot u_p \big)} \leq \overline{\textit{lcg}}, \qquad \forall p \in P \label{for:lcg} \\
\underline{\textit{vcg}} \leq \frac{\mathbf{1}^\top \big( \textit{vm} \odot u_p \big)}{\mathbf{1}^\top \big( w \odot u_p \big)} \leq \overline{\textit{vcg}}, \qquad \forall p \in P \label{for:vcg}
\end{align}
where $\textit{lm} = \textit{ld} \otimes w$ and $\textit{vm} = \textit{vd} \otimes w$ are longitudinal and vertical moments, with $ w \in \mathbb{R}^{|K|\times|\textit{TR}|}_{>0}$ being container weights, and $\textit{ld}, \textit{vd} \in \mathbb{R}^{|B| \times |D|}$ denoting the respective longitudinal and vertical distances from the center of gravity.

The problem's scale and complexity, combined with data scarcity and a practical runtime limit of 10 minutes, make most models focus on constraint subsets for tractability. This highlights the need for efficient solution methods.

\section{Related Work}
Container stowage planning has been addressed via exact methods \cite{robertiDecompositionMethodFinding2018}, heuristics  \cite{changSolvingIntegratedProblem2022}, and hybrid frameworks \cite{bilicanMathematicalModelTwoStage2020}. Our recent survey notes that scalable methods for representative stowage planning and the MPP remain open \cite{vantwillerLiteratureSurveyContainer2024}. Much of the literature further focuses on deterministic, cost-minimizing variants, while practice often requires profit maximization under uncertain demand. We target this gap by optimizing the MPP under demand uncertainty.

Uncertainty in optimization is classically handled by stochastic programming \cite{powellReinforcementLearningStochastic2022}, which represents uncertainty via a scenario tree of size $\mathcal{O}(b^T)$ with branching factor $b$ and stages $T$. This quickly becomes impractical for multi-stage decisions. Common mitigations include scenario reduction \cite{romischScenarioReductionTechniques2009}, decomposition (Benders/Lagrangian) \cite{rahmanianiBendersDecompositionAlgorithm2017}, and progressive hedging for non-anticipativity \cite{bolandCombiningProgressiveHedging2018}. 
Approximate approaches can also improve tractability: stochastic dual dynamic programs rely on known distributions \cite{shapiroAnalysisStochasticDual2011}, while sample average approximation uses Monte Carlo samples to solve deterministic surrogates \cite{chenSampleAverageApproximation2022}. 

Sequential stochastic decisions are often modeled as MDPs and solved with RL, avoiding explicit scenario-tree enumeration but shifting complexity to sampling and function approximation \cite{powellReinforcementLearningStochastic2022}. Enforcing constraints throughout learning is also challenging. Constrained MDP methods use penalties \cite{zhangPenalizedProximalPolicy2022}, primal--dual updates \cite{dingNaturalPolicyGradient2020}, or soft barriers \cite{wangEnforcingHardConstraints2023}, but may yield weak feasibility guarantees. Action masks and safety shields \cite{alshiekhSafeReinforcementLearning2018} can break differentiability, hinder learning when few actions are feasible, and require solving complex feasibility subproblems.

Another line of work addresses (state-dependent) action constraints in ML and DRL. Implicit optimization layers and constrained policy improvement operators \cite{agrawalDifferentiableConvexOptimization2019,phamOptLayerPracticalConstrained2018,pmlr-v161-lin21b} can ensure feasibility but typically solve and differentiate through constrained optimization at each online decision, which can be costly. Alternatives include projection mechanisms and feasible parameterizations \cite{bhatiaResourceConstrainedDeep2019,sanketSolvingOnlineThreat2020,dontiDC3LearningMethod2021}, and recent benchmarks highlight tradeoffs among feasibility, stability, and computational overhead in action-constrained DRL \cite{Kasaura2023Benchmarking}. For stochastic policies, feasible projection alters the induced action density, which biases policy-gradient updates. Several projection-based RL methods ignore this effect \cite{dontiDC3LearningMethod2021,sanketSolvingOnlineThreat2020}. In contrast to costly approaches, we propose a fast differentiable feasibility operator for {state-dependent convex polyhedral} constraints with a Jacobian-based log-probability correction, allowing feasible and efficient gradient-based training.

\section{Markov Decision Processes}
We formulate the MPP as an MDP with action constraints. Intuitively, the MDP tracks what is loaded and needed at each port (state), chooses how to load (action), updates the vessel and demand per port (transition), and measures profit after every decision (reward). To improve learning efficiency, we decompose the MDP into a sequence of simpler decisions, thereby reducing the action space. We provide supporting details on the MDP and its decomposition in~\refSuppMDP.

\subsection{Markov Decision Process}
Let ${\mathcal{M}} = ({S}, {X}, \mathcal{T}, {\mathcal{R}}, {P}^{N_P\text{-}1}_1, \gamma)$ define an episodic discounted MDP, where ${S}$ is a set of states, ${X}$ is a set of actions, $ \mathcal{T}: {S} \times {X} \rightarrow \Delta({S})$ is the transition function with $\Delta({S})$ being a distribution over set $S$, $ {\mathcal{R}}: {S} \times {X} \times {P}^{N_P\text{-}1}_1 \rightarrow \mathbb{R}$ is the reward function, ${P}^{N_P\text{-}1}_1$ is the finite horizon of load ports with final port $N_P\text{-}1$, and $\gamma \in (0,1)$ is the discount factor. 

The state ${s}_p = (u_p, q_p, \zeta)$ includes vessel utilization $u_p \in  \mathbb{R}_{\geq0}^{n_u}$, realized demand $q_p \in  \mathbb{R}_{\geq0}^{n_q}$, and instance parameters $\zeta$. The initial state $s_0  = (u_0, q_0,\zeta)$ has an empty vessel $u_0 = \mathbf{0}^{n_u}$,  realized demand $q_0$ of first port, and a generated problem instance $\zeta$. Note that $n_u = |B|\times|D|\times|K|\times|\textit{TR}|$, $n_c = |B|\times|D|$, and $n_q = |K|\times|\textit{TR}|$ are the shapes of utilization, location, and demand, respectively. 

At each port $p$, an action $x_p \in \mathbb{R}_{\geq 0}^{n_u}$ specifies how many containers of each group to load at each vessel position. To ensure safety and efficiency, actions must satisfy a set of linear but coupled constraints defined by the polyhedron
\(
\textit{PH}(s_p) = \left\{ x_p \in \mathbb{R}^{n_u}_{\geq 0} : A(s_p)x_p \leq b(s_p) \right\},
\)
where $A(s_p) \in \mathbb{R}^{m_u \times n_u}$ is the constraint matrix and $b(s_p) \in \mathbb{R}^{m_u}$ is the vector of bounds, with $m_u$ constraints. 

The constraints below are expressed in action $x_p$ and pre-loading utilization $u'_p$, i.e., the utilization after discharging cargo. We use real-valued actions, as downstream planning (e.g., SPP) discretizes the plans~\cite{pacinoFastGenerationNearoptimal2011}. Constraint \eqref{for:fr_demand} limits $x_p$ by demand $q_p$, while Constraint \eqref{for:fr_capacity} limits $x_p$ by the available TEU capacity, with TEU vector $\mathit{teu}\in \{1,2\}^{n_q}_{>0}$, and TEU capacity $c \in \mathbb{Z}^{n_c}_{>0}$. Here, $\mathbf{1}_{n_c}$ contracts the $(B,D)$ axes to leave $n_q$, and $\textit{teu}$ contracts the $(K,\textit{TR})$ axes to leave $n_c$. Constraints \eqref{for:fr_lcg_lb}--\eqref{for:fr_lcg_ub} enforce lcg limits, while Constraints \eqref{for:fr_vcg_lb}--\eqref{for:fr_vcg_ub} impose similar bounds on vcg. These stability bounds are derived from Constraints \eqref{for:lcg} and \eqref{for:vcg}.
\begin{align}
\quad \mathbf{1}_{n_c} {x}_p & \leq \; q_p  \label{for:fr_demand}\\
\quad  {x}_p \textit{teu} & \leq \; c - u'_p \textit{teu} \label{for:fr_capacity}\\
\text{-}\mathbf{1}^\top \! \big((\textit{lm}\!-\!\underline{\textit{lcg}} w)\!\odot\!{x}_p \big) & \leq \! \text{-}\mathbf{1}^\top\! \big( (\underline{\textit{lcg}} w\!-\!\textit{lm})\!\odot\!u'_p \big) \label{for:fr_lcg_lb} \\
\;\mathbf{1}^\top \! \big( (\textit{lm}\!-\!\overline{\textit{lcg}} w)\!\odot\!{x}_p \big) & \leq \! \; \mathbf{1}^\top\!\big( (\overline{\textit{lcg}} w\!-\!\textit{lm})\!\odot\!u'_p \big) \label{for:fr_lcg_ub} \\
\text{-}\mathbf{1}^\top \! \big((\textit{vm}\!-\!\underline{\textit{vcg}} w)\!\odot\!{x}_p \big) & \leq \! \text{-}\mathbf{1}^\top\!\big( (\underline{\textit{vcg}} w\!-\!\textit{vm})\!\odot\!u'_p \big) \label{for:fr_vcg_lb} \\
\;\mathbf{1}^\top \! \big((\textit{vm}\!-\!\overline{\textit{vcg}} w)\!\odot\!{x}_p \big) & \leq \! \; \mathbf{1}^\top\!  \big( (\overline{\textit{vcg}} w\!-\!\textit{vm})\!\odot\!u'_p \big) \label{for:fr_vcg_ub}
\end{align}

A stochastic transition function \( \mathcal{T}(s_{p+1} \! \mid \! s_p, x_p) \) updates a state to port $p\text{+}1$: 
\begin{itemize}
    \item Upon port arrival, the true demand $q_{p+1}$ is revealed, while future port demand remains unknown.
    \item Cargo is unloaded: $u_{p+1} = u_p \odot (1 - \mathbf{e}_p^-)$, with $\mathbf{e}_p^-$ marking discharge cargo.
    \item New cargo is loaded: $u_{p+1} = u_{p+1} + x_p$, where $x_p$ should satisfy $\textit{PH}(s_p)$.
\end{itemize}

Equation~\eqref{for:reward} defines a reward function that computes profit as revenue minus costs for any state $s$, action $x$, and port $p$. Revenue is computed per transport--cargo class as the elementwise minimum between loaded containers and realized demand: $\textit{rev}^\top \min(x^\top\mathbf{1}_{n_c}, q)$, where $\textit{rev} \in \mathbb{R}^{n_q}_{>0}$ is the vector of per-class revenues, $x^\top \mathbf{1}_{n_c} \in \mathbb{R}^{n_q}_{\ge 0}$ aggregates loaded containers over locations $(B, D)$, and the $\min$ is taken elementwise so that containers exceeding demand $q$ earn no revenue. Costs are determined using two state-dependent auxiliary variables: hatch overstows $\textit{ho}(s, p) \in \mathbb{R}^{|B|}_{\ge 0}$ and excess crane moves $\textit{cm}(s, p) \in \mathbb{R}^{|B|-1}_{\ge 0}$. Coefficients weight both components $\textit{ct}^{\textit{ho}} \in \mathbb{R}_{>0}$ and $\textit{ct}^{\textit{cm}} \in \mathbb{R}_{>0}$, respectively.
\begin{align} \label{for:reward}
\mathcal{R}(s, x, p) = \textit{rev}^\top \min\!\big(x^\top \mathbf{1}_{n_c},\, q\big) - \textit{ct}^{\textit{ho}}\mathbf{1}^\top \textit{ho}(s, p)- \textit{ct}^{\textit{cm}}\mathbf{1}^\top \textit{cm}(s,p)
\end{align}

\subsection{Decomposed Markov Decision Process}
\begin{figure}[t!]
    \centering
    \includegraphics[width=\linewidth]{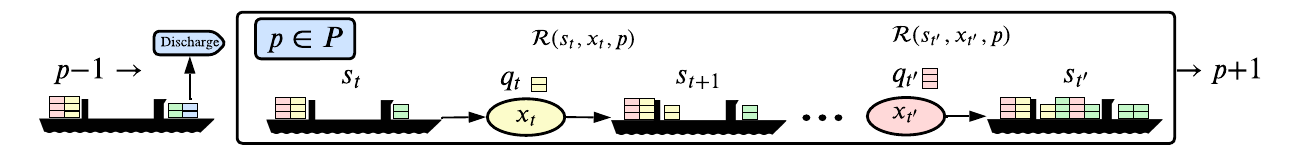}
    \caption{Illustration of the decomposed MDP for MPP. At port $p$, the vessel discharges cargo, observes realized demand $q_t$, then sequentially loads container type–POD pairs before sailing to the next port $p{+}1$. Colors indicate PODs.} 
\label{fig:mdp_illustration}
\end{figure}

The MDP has an action space of size $|{X}| \propto |B| \cdot |D| \cdot |K| \cdot |P|$, where each action ${x}_p$ determines how cargo types and transport options are placed on the vessel per port. However, this action space is large, which can hinder learning efficiency \cite{kanervistoActionSpaceShaping2020}. To address this, we decompose the MDP into granular, sequential steps based on an index $(i,j,k) \; \forall (i,j) \in \textit{TR}, k \in K$. Figure \ref{fig:mdp_illustration} illustrates the decomposed MDP for a single port. Instead of placing all transport and cargo types simultaneously, we take a decomposed action for each transport $(p,j)$ and cargo type $k$ at port $p$, then departing to a new port. This reduces the action space to $|X| = |B| \cdot |D|$, while unfolding transports and cargo types over an extended time horizon $t \in H = \{0,1,\dots,T_\textit{seq}\}$ with 
\(T_\textit{seq} = |K| \cdot |\textit{TR}|-1.\) 

The state $s_t = (u_t, q_t, \zeta)$ depends on time step $t$, where $u_t \in \mathbb{R}^{n_u}_{\geq 0}$ is vessel utilization, and $q_t \in \mathbb{R}^{n_q}_{\geq 0}$ is realized demand. The instance \( \zeta \) remains unchanged. Given the time $t$, however, we can extract relevant parameters from \( \zeta \), such as \( (\textit{pol}_t, \textit{pod}_t, k_t, \textit{rev}^{(\textit{pol}_t, \textit{pod}_t, k_t)}) \). 
Action $x_t \in \mathbb{R}^{n_c}_{\geq0}$ assigns real number of containers to utilization $u_t$ for step $t$. Each action is subject to $\textit{PH}(s_t) =  \{x_t \in \mathbb{R}^{n_c}_{\ge 0} : A(s_t) x_t \leq b(s_t)\} $. Here, $ A(s_t) \in \mathbb{R}^{m_c \times {n_c}} $ is the constraint matrix, $ b(s_t) \in \mathbb{R}^{m_c} $ is the bound vector, and $m_c$ is the number of constraints. Constraints \eqref{for:fr_demand}--\eqref{for:fr_vcg_ub} are re-indexed for the decomposed steps by replacing \((x_p,s_p)\) with \((x_t,s_t)\), which collapses $x_t$ onto the $(B,D)$ axes alone. The inequalities define the same polyhedron, but are instantiated at step $t$ based on \(s_t\), yielding \(\textit{PH}(s_t)\).
At each time \( t \), the transition includes loading, where \( x_t \) is added to \( u_t \). Discharges and demand realizations occur when arriving at new ports \( t \in T_{\text{new port}} \).
The revenue at step \( t \) is computed as \( \textit{rev}(\textit{pol}_t, \textit{pod}_t, k_t) \min(\mathbf{1}^\top x_t, q_t^{(\textit{pol}_t, \textit{pod}_t, k_t)}). \) Costs, however, require all discharge and loading operations at port $p$ to be finalized, as they depend on the vessel utilization $u_t$ before departure, aggregated at the final step \( t \in T_{\text{leave port}} \). The hatch-overstow and crane-move costs depend on the full vessel utilization at departure ($t \in T_{\text{leave port}}$), so they are defined at port granularity rather than per step. Decomposing them into per-step signals would require approximations. We therefore evaluate the cost signal once per port $p$, which yields structurally sparse costs.

\section{Proposed Architecture} \label{sec:arch}
\begin{figure}[b!]
    \centering
    \includegraphics[width=0.65\linewidth]{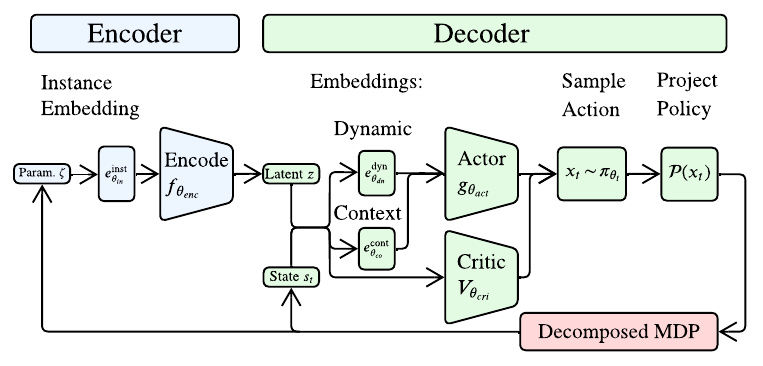}
    \caption{DRL architecture with feasibility projection for actor-critic methods.} \label{fig:drl_architecture}
\end{figure}

Figure~\ref{fig:drl_architecture} shows our projected encoder--decoder model, which embeds instance, context, and dynamic data to parameterize the multivariate Gaussian policy
\(
\pi_\theta(\cdot \mid s_t) = \mathcal{N}\!\left(\mu_\theta(s_t), \operatorname{diag}(\sigma_\theta^2(s_t))\right).
\)
A sampled action vector \(x_t \sim \pi_\theta(\cdot \mid s_t)\) is projected by layer \(\mathcal{P}\) to ensure feasibility, and the adjusted action is applied in the decomposed MDP. We train the projected policy using actor--critic RL, with the critic evaluating actions along the trajectory.

\begin{figure}[b!]
    \centering
    \includegraphics[width=0.65\linewidth]{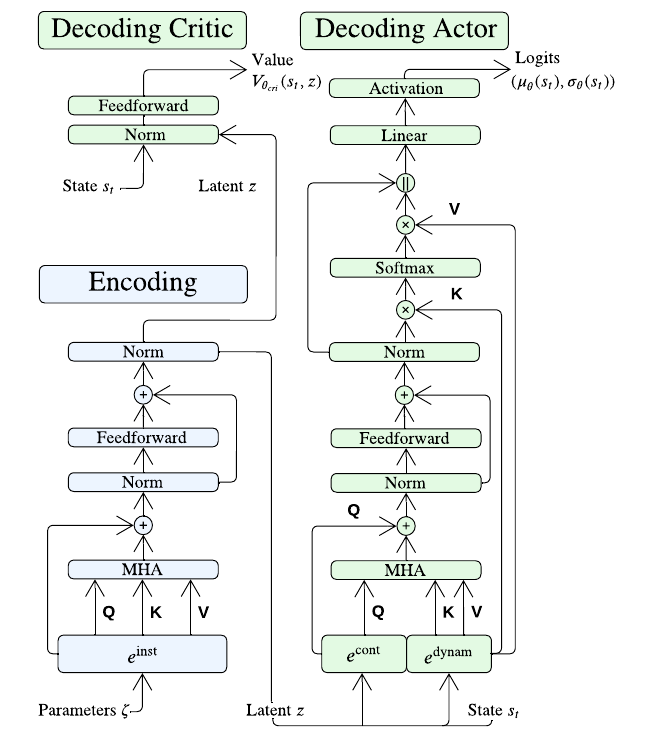}
    \caption{Layers of the encoder and the actor-critic decoder.}
    \label{fig:encoder_decoder_models}
\end{figure}
 
\subsection{Encoder-Decoder Model}
Our model, inspired by \cite{koolAttentionLearnSolve2019}, is illustrated in Figure \ref{fig:encoder_decoder_models}.
The instance embedding \( e^\text{inst}(\zeta) \), parameterized by \( \theta_\textit{in} \), maps problem instance $\zeta$ to a feature representation used by the encoder to condition policies on instances. Sinusoidal positional encoding is then applied to inform the model of element order in sequences \cite{vaswaniAttentionAllYou2017}.
An attention encoder $f(e^\text{inst}(\zeta))$ parameterized by ${\theta_\textit{enc}}$  maps embedding $e^\text{inst}(\zeta)$ to latent variable $z$ using multi-head attention (MHA) with keys (K), queries (Q) and values (V) to identify relevant features dynamically and handle variable input size \cite{vaswaniAttentionAllYou2017}.  Then, we use a feed-forward network (FFN) with ReLU activation, layer normalization, residual connections, and dropout. 
The context embedding \( e^\text{cont}(u_t, z) \), parameterized by \( \theta_\textit{co} \), extracts features from the current solution \( u_t \) and the latent variable \( z \). This embedding provides the MHA query, helping the attention mechanism focus on relevant information, with the solution context at time $t$, enabling the policy to make informed decisions on the current state.
The dynamic embedding \( e^\text{dyn}(q_t, z) \), parameterized by \( \theta_\textit{dn} \), extracts features from demand \( q_t \) across all time steps in the horizon \( H \), capturing temporal patterns over the episode. By combining actual demand with latent information, attention selects (key) and retrieves (values) demand-relevant context over time, improving anticipation of future conditions and long-term decisions.

Following \cite{koolAttentionLearnSolve2019}, the actor decoder $g(e^\text{cont}(u_t, z), e^\text{dyn}(q_t, z))$ is an attention model (AM) parameterized by $\theta_\textit{act}$. The MHA layer uses context embedding as queries and dynamic embedding as keys and values encoding demand across the horizon, with a forward-looking mask on steps $\{t, t\!+\!1, \dots, T_\textit{seq}\}$ to link present context with future trends. An FFN extracts features, a pointer mechanism selects relevant future steps \cite{vinyalsPointerNetworks2015}, and softplus activation then outputs positive parameters $({\mu}_\theta(s_t), {\sigma}_\theta(s_t))$.
Our critic model $V(s_t,z)$, parameterized by $\theta_\textit{cri}$, estimates the value of state $s_t$ and latent variable $z$ through an FFN outputting $V_{\theta_\textit{cri}}(s_t, z) \in \mathbb{R}$.
The actor logits parameterize a multivariate Gaussian policy
\(
\pi_\theta(\cdot \mid s_t) = \mathcal{N}\!\left(\mu_\theta(s_t), \operatorname{diag}(\sigma_\theta^2(s_t))\right),
\)
from which a raw action vector is sampled as $x_t \sim \pi_\theta(\cdot \mid s_t)$ and then used to construct solutions.

\subsection{Feasibility Regularization in Actor-Critic Methods}
Actor-critic methods such as proximal policy optimization (PPO) ~\cite{schulmanProximalPolicyOptimization2017} and soft actor critic (SAC) ~\cite{haarnojaSoftActorCriticOffPolicy2018} are widely used in unconstrained problems where feasibility is enforced implicitly through the environment dynamics. A common approach for handling constraints without environment enforcement is to augment the actor loss with a soft penalty term \cite{zhangPenalizedProximalPolicy2022}. This actor loss is defined as:
\begin{align} 
\mathcal{L}(\theta) &= \mathcal{L}_{\text{actor}}(\theta) +  \mathcal{L}_{\text{feas}}(\theta)  \label{for:actor_loss} \\
\mathcal{L}_{\text{feas}}(\theta) &= \mathbb{E}_t \left[ \lambda_{{f}}^\top\left( A(s_t) x_\theta(s_t) - b(s_t) \right)_{+} \right] \label{for:feas_loss}
\end{align}
where \( ( \cdot )_{+}\) denotes the ReLU function applied elementwise, penalizing only constraint violations. The vector \( \lambda_f \in \mathbb{R}_{>0}^m \) weights the regularization terms for the \(m\) constraints. For algorithms that do not propagate gradients through the action, such as PPO, \(x_\theta(s_t)\) is replaced by the policy mean \( \mu_\theta(s_t) \). Tuning \( \lambda_f \) is difficult under coupled constraints and even more so in dynamic, state-dependent feasible regions \( \textit{PH}(s_t) \), where the reward--feasibility trade-off varies over time~\cite{dontiDC3LearningMethod2021}. Primal--dual methods learn the policy and \( \lambda_f \) jointly, but can still struggle to enforce hard constraints. Both methods are mainly useful as constrained RL baselines.

\subsection{Feasibility Projection Layers}
Rather than penalizing violations after they occur, we ensure feasibility by mapping each sampled action onto the feasible region before it is used in the MDP or the policy update. To avoid feasibility-loss augmentation, we insert a differentiable feasibility map $\mathcal{P}$ that transforms sampled actions $x_t\sim\pi_\theta(\cdot\mid s_t)$ into constraint-satisfying actions before they enter the decomposed MDP or the policy update. For the state-dependent convex polyhedron $\textit{PH}(s_t)$, we execute $x'_t=\mathcal{P}(x_t)\in\textit{PH}(s_t)$. Because $\mathcal{P}$ changes the action distribution, the policy gradient would be biased unless this transformation is accounted for. We therefore apply a change-of-variables log-probability correction using the map's Jacobian~\cite{bishopPatternRecognitionMachine2006}, valid wherever \(\mathcal{P}\) is locally invertible, which holds almost everywhere as \(\mathcal{P}\) is piecewise-linear. This keeps the gradient estimates unbiased (a.e.), hence the name UVP. Unlike implicit differentiable convex optimization layers with intractable Jacobians and biased, costly gradients~\cite{agrawalDifferentiableConvexOptimization2019}, our feasibility maps admit explicit Jacobians, making the correction efficient. 

Algorithm~\ref{alg:viol_projection} implements the unbiased violation projection (UVP) by iteratively reducing constraint violation with respect to $\textit{PH}(s_t)$. Starting from a raw action $x$, UVP computes the violation $\mathcal{V}(x)=(Ax-b)_+$ and updates $x\leftarrow x-\eta_v A^\top\mathcal{V}(x)$ to reduce the total violation. The update corresponds to gradient descent on the squared violation $\|(Ax-b)_+\|_2^2$. Repeating this update for a fixed number of steps (training) or until violation improvement is below a threshold (inference) yields a (near-)feasible action while remaining differentiable almost everywhere, enabling end-to-end learning with the Jacobian log-probability adjustment. Since this objective is $2\|A\|_2^2$-smooth and can be steep when $A$ has large singular values, we normalize the step size as $\eta_v=\tilde{\eta}_v/\|A\|_2^2$ to ensure stable iterations. Technical details on the Jacobian correction and UVP convergence are provided in~\refSuppFeas.

\begin{algorithm}[t!]
\scriptsize
\caption{Unbiased violation projection (UVP) layer}
\label{alg:viol_projection}
\begin{algorithmic}[1]
\REQUIRE $x \in \mathbb{R}^{n}_{\ge 0}$, parameters $(A, b, \eta_v, \delta_v)$, iterations
\STATE Initialize $x' \gets x$
\STATE Define $\mathcal{V}(x) \gets (Ax - b)_{+}$
\FOR{$i = 1$ \textbf{to} iterations}
    \STATE Set $x \gets x'$
    \STATE Update $x' \gets x - \eta_v A^\top \mathcal{V}(x)$
    \IF{$|\mathbf{1}^\top \mathcal{V}(x') - \mathbf{1}^\top \mathcal{V}(x)| \leq \delta_v$}
        \STATE \textbf{break}
    \ENDIF
\ENDFOR
\RETURN $x'$
\end{algorithmic}
\end{algorithm}

\section{Experimental Results} \label{sec:exp_res}
We compare methods across demand distributions on objective value (Obj.) and training and test times in seconds (Time). An ablation study assesses the contribution of individual components, while managerial insights analyze information value, computational cost, adapting to uncertainty, and scaling to larger instances. DRL used an NVIDIA RTX A6000 GPU, while CPLEX used an AMD EPYC 7742 48-core CPU. Hyperparameters are consistent across DRL models.

\subsection{Experimental Setup}
Training instances are based on 4-port voyages and a 1,000 TEU vessel. These synthetic instances are sampled from Gaussian distributions $\mathcal{N}(\mu^{(i,j,k)}, \sigma^{(i,j,k)})$ for all $(i,j)\in\textit{TR}$ and $k\in K$, where means $\mu^{(i,j,k)}$ are uniformly randomized and the standard deviations follow $\sigma^{(i,j,k)} = \textit{CV}\mu^{(i,j,k)}$ with coefficient of variation $\textit{CV}$. While instances are synthetic, the distributional parameters follow established instance generators from prior work \cite{avrielStowagePlanningContainer1998,dingStowagePlanningContainer2015}, ensuring a realistic transport mix and high capacity utilization. Generalization instances use continuous uniform samples $\mathcal{U}(\textit{lb}^{(i,j,k)}, \textit{ub}^{(i,j,k)})$ for all $(i,j)\in\textit{TR}$ and $k\in K$, where  (\textit{lb}, \textit{ub}) are selected to match the spread of the training distribution, isolating effects of distributional shift. Details are found in~\refSuppGen.

Guided by a taxonomy of action-constrained DRL \cite{Kasaura2023Benchmarking} and loss-augmented RL \cite{dingNaturalPolicyGradient2020,zhangPenalizedProximalPolicy2022}, we include one representative baseline from each major family. Our end-to-end RL baselines are penalty loss (Pen-DRL), a primal--dual approach (Lag-DRL), an implicit convex optimization layer (CP) \cite{agrawalDifferentiableConvexOptimization2019}, Frank--Wolfe policy improvement \cite{pmlr-v161-lin21b}, and $\alpha$-map projection \cite{sanketSolvingOnlineThreat2020}. For baselines that may output infeasible actions, we include an inference-time recovery layer (R) that projects actions exactly onto $\textit{PH}(s_t)$ via CP. We use on-policy PPO \cite{schulmanProximalPolicyOptimization2017} and off-policy SAC \cite{haarnojaSoftActorCriticOffPolicy2018} to show that trends are agnostic to DRL algorithms. As a non-learning baseline, we include a multi-stage stochastic program on an explicit scenario tree (SP) with the same information structure as DRL. We exclude methods that optimize alternative uncertainty objectives (e.g., robust/deterministic methods) and feasibility mechanisms that are non-differentiable or enforce subsets of the constraints, as these are not compatible with gradient-based action-constrained RL. \refSuppSP ~and~\refSuppDRL ~provide implementation details of the SP and RL, respectively.

\subsection{Policy Evaluation}
Table~\ref{tab:comparison} reports the performance of methods that consistently produce feasible solutions. Our AM-P with UVP+R achieves the best average performance, with variability comparable to the AM-P baselines (CP, FW, $\alpha$). Its objective value is significantly higher than loss-augmented DRL and within about 10\% of the SP upper bound, while matching the training and inference time of loss-augmented DRL. In contrast, SP has an average runtime of 52–57 minutes. Among AM-P models, UVP achieves objectives similar to CP and FW, while $\alpha$ performs slightly worse. The AM-P baselines also require much higher training cost than UVP, whereas the inference cost is small for each AM-P. These trends are consistent across test and generalization instances, and across off-policy (SAC) and on-policy (PPO) settings. Overall, AM-P with UVP+R achieves the best trade-off across baselines, delivering high-quality solutions for the MPP under demand uncertainty at substantially lower training cost.

\input{tables/results_table}

Figure~\ref{fig:ablation} presents an ablation study isolating the contribution of key components; Figures~\ref{fig:ablation}(b) and \ref{fig:ablation}(c) contain overlapping lines with similar values. 

In Figure~\ref{fig:ablation}(a), high-quality solutions lie in the top-left corner. Compared to UVP, feasibility recovery (+R) and a larger budget of 150 iterations (UVP*) both reduce violations, though UVP* requires tuning the iteration budget. Even so, UVP, UVP+R, and UVP* reach similar profits despite residual training violations, with small violation norms of about 0.07, 0.004, and 0.003, respectively. Replacing projection with loss-augmentation (Pen, Lag) degrades the objective and feasibility. Removing projections (e.g., \st{UVP}) shows that AM-P relies on them to find high-quality solutions. 

Figure~\ref{fig:ablation}(b) plots the average training objective with 95\% confidence intervals for SAC and PPO over five random seeds. With Jacobian correction (JC), training is more stable; without it (NC), runs often experience policy collapse. The Jacobian adjustment is irrelevant for PPO as the change-of-variable term is an additive correction, $\log|\det J(a,s)|$, independent of $\theta$. Since PPO uses the likelihood ratio $\log \pi_\theta(a\mid s)-\log \pi_{\theta_{\mathrm{old}}}(a\mid s)$, this term cancels when applied consistently, leaving the updates unchanged. Overall, PPO training remains stable. Despite the sparse per-port cost signal, both PPO and SAC with JC converge stably over multiple seeds, indicating the signal is sufficient for the critic to learn effective policies under structurally sparse costs. 

Figures~\ref{fig:ablation}(c) and ~\ref{fig:ablation}(d) show the effect of UVP iterations without feasibility recovery. Beyond 100--150 iterations, SAC and PPO with normalized ${\eta}_v$ show diminishing returns and higher compute costs. Compared with static $\tilde{\eta}_v$, normalized $\eta_v$ avoids parameter tuning and reaches high-profit, low-violation solutions faster and more reliably. 

\input{plots/ablations}  

\subsection{Managerial Insights}
Figure \ref{fig:comparison_subfigures} performs sensitivity analyses on SP scenario size, demand uncertainty, and voyage length, reporting averages and standard deviations with $N=30$.

Figure~\ref{fig:comparison_subfigures}(a) reports profits of SP and our AM-P policies. DRL shares SP's information structure, as it observes the realized demand per port and uses the same demand model. As the number of scenarios increases, SP better approximates the uncertainty and its profit remains stable. At 32 scenarios, DRL attains about 90\% of the SP profit, showing that learning can approach the expected upper bound, with a remaining gap of about 10\%.

Figure~\ref{fig:comparison_subfigures}(b) shows the average computational cost. SP’s time grows substantially with the number of scenarios, from 4 seconds at 4 scenarios to 52 minutes per instance at 32 scenarios, exceeding the practical limit of 10 minutes beyond 16 scenarios. Using SAC and PPO, we train AM-P with UVP within 25-30 minutes. Inference, however, is near-instant for all DRL methods. When amortizing training time over $N=30$, SAC and PPO cost 45–60 seconds per solution. Hence, DRL is efficient and meets the practical limit of 10 minutes.

Figure~\ref{fig:comparison_subfigures}(c) shows how policies adapt as demand variability measured by the coefficient of variation (CV) changes across unseen instances. Our samples are clamped at 0 to ensure non-negativity, which can skew demand distributions at higher CVs. As variability and uncertainty increase, SAC and PPO policies exhibit stable, feasible performance and slightly increase profit by exploiting the demand skewness, demonstrating their adaptability to shifting distributions.

Figure~\ref{fig:comparison_subfigures}(d) shows that policies trained in $|P|=4$ can scale to longer voyages $|P|\in\{6,8,10\}$, and hence policies perform a zero-shot transfer without retraining. However, the policy’s share of the maximum revenue declines as $|P|$ grows, alongside an increase in per-step average capacity utilization from 55\% to 80\%. This suggests that tighter capacity leaves less flexibility to load demand. Inference costs grow linearly with the problem size, while solutions remain feasible. Note that scaling to larger vessels currently requires retraining.

\input{plots/sensitivity_analysis}

\subsection{Industrial Realism and Scope}
Large-scale stowage datasets are hard to obtain as demand, contractual, and planning data are proprietary and fragmented across systems. We therefore use expert-informed generators with realistic operational parameters and high capacity utilization to stress the interaction between uncertainty and constraints. This supports reproducible evaluation in realistic conditions, but does not fully capture industrial dynamics, non-convex constraints, or downstream coordination. We thus view the results as evidence of deployment potential, while industrial deployment will require representative generators at scale or large-scale datasets.

\section{Conclusion and Future Directions}
We presented a DRL approach for the MPP under demand uncertainty, modeling the problem as a stochastic MDP with hard state-dependent constraints. Our encoder–decoder attention policy with differentiable feasibility projections enables the learning of feasible policies under limited compute budgets. The resulting policies provide fast, feasible decision support for the MPP within practical runtime limits, while remaining robust under distribution shift and scaling to longer voyages. More broadly, we show that differentiable feasibility mechanisms can make DRL viable for realistic constrained planning. Future work should address efficient non-convex projections, validation on representative industrial-scale instances, and integration with related planning tasks.

\begin{credits}
\section*{Ethical Statement}
Human oversight is critical in {container shipping} decision-support tools for fairness, accountability, and transparency. While models enhance decision-making and provide optimized, feasible plans, human judgment remains essential, especially in safety-critical situations like vessel stability and dangerous cargo. Optimizing solely for revenue might inadvertently bias cargo prioritization, impacting customers and voyages. Safeguards such as challengeable outputs and continuous auditing help mitigate automation bias and build trust in these systems. Efficient vessel utilization also reduces fuel consumption and greenhouse gas emissions.

\section*{Generative AI Statement} 
During the preparation of this manuscript, the authors used generative AI tools to assist with language editing and to support code development. All content was subsequently reviewed and verified by the authors, who take full responsibility for the final result.

\section*{Acknowledgments}
This work is partially sponsored by the IT University of Copenhagen and the Danish Maritime Fund under research grant 2025-039 and grant 2023-047 for the Maritime Hub at the IT University of Copenhagen.

\end{credits}
%
%
%

\bibliographystyle{splncs04}
\bibliography{references/references.bib} 

\clearpage
\appendix

\chapter*{Supplementary Material for Navigating Demand Uncertainty in Container Shipping: Deep Reinforcement Learning for Enabling Adaptive and Feasible Master Stowage Planning}

\section{MDP Formulation of the Master Planning Problem}\label{app:MDP}

\subsection{Index sets and instance parameters}\label{app:sets_params}
Let $P=\{1,\dots,N_P\}$ be the ordered ports and $\textit{TR}\subseteq P^2$ the transport pairs $(i,j)$ with $i<j$.
For a port $p\in P$, define
\[
\textit{TR}^{\textit{OB}}_p=\{(i,j)\in\textit{TR}\mid i\le p<j\},\quad
\textit{TR}^{\textit{ROB}}_p=\{(i,j)\in\textit{TR}\mid i<p<j\},
\]
\[
\textit{TR}^{+}_p=\{(p,j)\in\textit{TR}\mid j>p\},\quad
\textit{TR}^{-}_p=\{(i,p)\in\textit{TR}\mid i<p\},\quad
\textit{TR}^{M}_p=\textit{TR}^{+}_p\cup \textit{TR}^{-}_p.
\]

Let vessel locations be indexed by $(b,d)\in B\times D$ with $n_c:=|B||D|$, and let transport--cargo classes be indexed by $((i,j),k)\in \textit{TR}\times K$ with $n_q:=|\textit{TR}||K|$.
We represent utilization as a nonnegative tensor
\[
u=\big(u^{(b,d,(i,j),k)}\big)\in\mathbb{R}_{\ge0}^{n_c\times n_q},
\]
and write $\mathrm{vec}(u)\in\mathbb{R}^{n_u}_{\ge0}$ for its vectorization with $n_u=n_c n_q$. For each class $((i,j),k)\in \textit{TR}\times K$, the instance parameter is $\zeta^{(i,j,k)}=(\mu^{(i,j,k)},\sigma^{(i,j,k)},\textit{teu}^{(k)},w^{(k)},\textit{rev}^{(i,j,k)})$. Demand is sampled as $q^{(i,j,k)}\sim\mathcal{Q}(\mu^{(i,j,k)},\sigma^{(i,j,k)})$. We define the following quantities, where the parameter instantiations are given in \refSuppDRL:
\begin{align*}
\textit{teu}(k)= & \; \mathbb{I}\{\kappa_1=\text{20 ft.}\}+2\,\mathbb{I}\{\kappa_1=\text{40 ft.}\}, \\
w(k)=  & \; \mathbb{I}\{\kappa_2=\text{Light}\}+2\mathbb{I}\{\kappa_2=\text{Medium}\}+3\mathbb{I}\{\kappa_2=\text{Heavy}\}, \\
\textit{rev}(i,j,k)= & \; (j-i)+0.1-\textit{LR}(j-i)\,\mathbb{I}\{\kappa_3=\text{Long}\},
\end{align*}

\subsection{MDP state, action, and transition}\label{app:mdp}
We consider a sequential decision process over $T_{\textit{seq}}:=|\textit{TR}||K|$ steps that iterates through tuples
$(\textit{pol}_t,\textit{pod}_t,k_t)$ with $\textit{pol}_t<\textit{pod}_t$.
The state is
\[
s_t=(u_t,\,q_t), \qquad u_t\in\mathbb{R}^{n_c\times n_q}_{\ge0},\; q_t\in\mathbb{R}^{n_q}_{\ge0}.
\]
At arrival to a load port, demand for $\textit{TR}^+_{\textit{pol}_t}$ is revealed (all other components may be zero-masked).

Actions allocate the load of the current class across vessel locations \(x_t\in\mathbb{R}^{n_c}_{\ge0}.\) Let $e_t^{+}\in\{0,1\}^{n_q}$ select the current class $((\textit{pol}_t,\textit{pod}_t),k_t)$ and let $e_t^{-}\in\{0,1\}^{n_q}$ select classes discharged at the current port. The transition applies (i) discharge at new-port steps, then (ii) load:
\[
u_{t+1}=u_t\odot(1-e_t^{-}) + x_t\otimes e_t^{+},
\]
where $x_t\otimes e_t^{+}\in\mathbb{R}^{n_c\times n_q}$ places $x_t$ in the selected class and zeros elsewhere.

\paragraph{Port-change steps.}
Let $\Psi(p):=|K|\sum_{r=1}^{p-1}(N_P-r)$ be the number of steps completed before load port $p$.
Then $T_{\text{new port}}=\{\Psi(p)\mid p=2,\dots,N_P-1\}$ and
$T_{\text{leave port}}=\{\Psi(p+1)-1\mid p=1,\dots,N_P-2\}$.

\subsection{Feasible region and constraint reduction}\label{app:FR_time}
Let the original MPP constraints be $A'\,\mathrm{vec}(u)\le b'$ with $u\ge0$.
Write $u_t=u'_t+u_t^{+}$ where $u'_t$ is the pre-load utilization after discharges and $u_t^{+}$ is the load increment.
Then feasibility is
\[
A'\,\mathrm{vec}(u_t^{+}) \le b' - A'\,\mathrm{vec}(u'_t), \qquad u_t^{+}\ge0.
\]
At step $t$, the load increment affects only the current class, i.e., $u_t^{+}=x_t\otimes e_t^{+}$, yielding a reduced constraint set
\[
A(s_t)x_t\le b(s_t), \quad x_t\ge0,
\]
where $A(s_t)$ and $b(s_t)$ are obtained by selecting the columns of $A'$ corresponding to the current class and by shifting the RHS by $A'\mathrm{vec}(u'_t)$.

\subsection{MPP constraints per decomposed step}\label{app:mpp_constraints}
Let $q_t$ denote the demand for the current class, and let $\textit{teu}_t$ and $w_t$ denote the TEU and weight of the current cargo type. The per-step MPP constraints are:
\begin{align*}
\text{Demand:}\qquad 
&\mathbf{1}^\top x_t \le q_t,\\[2mm]
\text{Capacity:}\qquad 
&\textit{teu}_t \mathbf{1}^\top x_t \le c - \textit{teu}^\top u'_t,
\end{align*}
where $\textit{teu}^\top u'_t$ is the TEU capacity occupied by onboard cargo before loading.

Let $\textit{lm}$ denote the longitudinal moment arm, broadcast over cargo classes, and recall that $u_t=u'_t+x_t\otimes e_t^{+}$. Then the upper-LCG bound can be written in linear form as
\begin{align*}
\frac{\mathbf{1}^\top (\textit{lm} \odot u_t)}{\mathbf{1}^\top (w \odot u_t)}
\le \overline{\textit{lcg}}
&\iff
\mathbf{1}^\top \big((\textit{lm}-\overline{\textit{lcg}}\,w)\odot u_t\big)\le 0 \\
&\iff
\mathbf{1}^\top \big((\textit{lm}-\overline{\textit{lcg}}\,w)\odot (x_t \otimes e_t^{+})\big)
\le
\mathbf{1}^\top \big((\overline{\textit{lcg}}\,w-\textit{lm})\odot u'_t\big).
\end{align*}
Analogous linear reformulations are obtained for the lower and upper bounds on both LCG and VCG.

\subsection{Auxiliary port-operation variables}\label{app:aux_vars}
Let $D^{\textit{below}}\subset D$ and $D^{\textit{above}}\subset D$ denote below- and on-deck locations.

\paragraph{Hatch movements and overstowage.}\hspace{1em}
For each bay $b$ and port $p$,
\begin{align*}
\textit{hm}_b(s,p)&=\mathbb{I}\Big\{\sum_{d\in D^{\textit{below}}}\sum_{(i,j)\in \textit{TR}^M_p}\sum_{k\in K} u^{(b,d,(i,j),k)} > 0\Big\}, \\
\textit{ho}_b(s,p)&=\textit{hm}_b(s,p)\sum_{d\in D^{\textit{above}}}\sum_{(i,j)\in \textit{TR}^{\textit{ROB}}_p}\sum_{k\in K} u^{(b,d,(i,j),k)}.
\end{align*}

\paragraph{Crane move imbalance.}\hspace{1em}
Define the per-port target
\[
\overline{\textit{cm}}(s,p)=(1+\delta^{\textit{cm}})\frac{2}{|B|}\sum_{(i,j)\in \textit{TR}^M_p}\sum_{k\in K} q_t^{(i,j,k)}.
\]
For adjacent bays $(b,b+1)$, excess moves are
\[
\textit{cm}_b(s,p)=\Bigg[\sum_{d\in D}\sum_{(i,j)\in \textit{TR}^M_p}\sum_{k\in K}\big(u^{(b,d,(i,j),k)}+u^{(b+1,d,(i,j),k)}\big)-\overline{\textit{cm}}(s,p)\Bigg]_+ .
\]

\section{Feasibility Mechanisms}\label{app:feas_proj}
We execute a transformed action obtained by post-processing a policy sample. This changes the executed-action distribution, so log-probabilities are corrected via change-of-variables (CoV) wherever the transformation is differentiable and locally invertible almost everywhere (a.e.).

\subsection{Log-probability under executed-action maps}\label{app:log_prob}
Let \(x_0\sim\pi_\theta(\cdot\mid s)\) and execute \(x=f(x_0)\) with \(f:\mathbb{R}^n\to\mathbb{R}^n\). On points where \(f\) is differentiable and locally invertible with \(\det J_f(x_0)\neq 0\), 
\[
\log \pi^f_\theta(x\mid s)=\log \pi_\theta(x_0\mid s)-\log\!\big|\det J_f(x_0)\big|,\qquad x=f(x_0).
\]
If \(f\) is not globally one-to-one, the exact pushforward density sums over all preimages \(x_0\in f^{-1}(x)\). We use the Jacobian correction only on the set where \(f\) is a.e. locally one-to-one.

\subsection{Unbiased Violation Projection (UVP)}\label{app:uvp}
We consider linear constraints \(Ax\le b\) with \(A\in\mathbb{R}^{m\times n}\), \(b\in\mathbb{R}^m\), and (optionally) nonnegativity \(x\in\mathbb{R}^n_{\ge 0}\). The feasible set is the polyhedron \(\textit{PH}=\{x\in\mathbb{R}^n_{\ge 0}: Ax\le b\}. \) We define the elementwise positive part \(((u)_+)_i=\max(0,u_i)\) and violation vector \(\mathcal V(x):=(Ax-b)_+\in\mathbb{R}^m .\)

\paragraph{One UVP step.}
Given \(\eta_v>0\), UVP applies
\[
\mathcal P(x):=x-\eta_v A^\top \mathcal V(x).
\]

\paragraph{Jacobian and local CoV admissibility.}
\(\mathcal P\) is differentiable at any \(x\) with \((Ax-b)_i\neq 0\) for all \(i\). Let \(\mathbf 1_{Ax-b>0}\in\{0,1\}^m\) be the indicator of violated constraints and \(D(x)=\mathrm{diag}(\mathbf 1_{Ax-b>0})\). Then
\[
J_{\mathcal P}(x)=\frac{\partial \mathcal P(x)}{\partial x}=I_n-\eta_v A^\top D(x)A,
\]
where \(I_n\) is the \(n\times n\) identity. A sufficient condition for \(J_{\mathcal P}(x)\) to be nonsingular is \(\eta_v\,\lambda_{\max}(A^\top D(x)A)<1 \), where $\lambda_{\max}$ is the largest eigenvalue. A uniform sufficient choice (valid for any active set) is \(\eta_v<\frac{1}{\|A\|_2^2}, \) as \(\lambda_{\max}(A^\top D(x)A)\le \lambda_{\max}(A^\top A)=\|A\|_2^2\), and \(\|A\|_2\) is the spectral norm. \\

\noindent\textbf{Caveat.} Since \(\mathcal P\) is piecewise-defined, it need not be globally one-to-one. The Jacobian correction \(-\log|\det J_{\mathcal P}|\) yields the correct local CoV term only on the (a.e.) set where \(\mathcal P\) is differentiable and locally invertible. If multiple distinct preimages map to the same output, the exact pushforward density would require summing over all preimages.

\noindent\textbf{Intuition.}
With a fixed violated set, \(J_{\mathcal P}(x)=I-\eta_v A^\top D(x)A\). The matrix \(A^\top D(x)A\) captures how strongly UVP corrects different directions. Along an eigenvector with eigenvalue \(\lambda\), local changes are scaled by \(1-\eta_v\lambda\). If \(\eta_v\lambda=1\), the Jacobian is singular (\(\det J_{\mathcal P}(x)=0\)) and CoV is undefined. Thus, it suffices to require \(\eta_v\,\lambda_{\max}(A^\top D(x)A)<1\). Since \(0\preceq D(x)\preceq I\), \(\lambda_{\max}(A^\top D(x)A)\le \|A\|_2^2\), giving the uniform bound \(\eta_v<1/\|A\|_2^2\).


\paragraph{\(K\)-step rollout.}
We define \(x_{k+1}=\mathcal P(x_k)\) for \(k=0,\dots,K-1\) and \(x_K=\mathcal P^{(K)}(x_0)\). For a.e.\ \(x_0\) where all Jacobians exist, \(\log\!\big|\det J_{\mathcal P^{(K)}}(x_0)\big| =\sum_{k=0}^{K-1}\log\!\big|\det J_{\mathcal P}(x_k)\big|. \) On the (a.e.) set where \(\mathcal P^{(K)}\) is differentiable and locally one-to-one,
\[
\log \pi^{(K)}_\theta(x_K\mid s)
=\log \pi_\theta(x_0\mid s)-\log\!\big|\det J_{\mathcal P^{(K)}}(x_0)\big|.
\]

\paragraph{Lipschitz constant and choosing \(\eta_v\).}
Where differentiable, \(\nabla g(x)=2A^\top(Ax-b)_+\). Since \(z\mapsto z_+\) is nonexpansive,
\[
\|\nabla g(x)-\nabla g(y)\|
\le 2\|A\|_2^2\|x-y\|,
\]
so \(\nabla g\) is \(L\)-Lipschitz with \(L\le 2\|A\|_2^2\). In practice, we set \(\eta_v=\upsilon/\widehat L\) with \(\upsilon\in(0,1)\) and \(\widehat L = \|A\|_2^2\). We use singular value decomposition to determine $\|A\|_2^2$. This yields \(\eta_v<2/\|A\|_2^2\le 4/L\).

\subsection{Violation descent guarantee}\label{app:uvp_proof}
We define \(g(x):=\|\mathcal V(x)\|_2^2\). On any region with a fixed active set (i.e., \(D(x)\) constant),
\(g\) is a smooth quadratic and \(\nabla g(x)=2A^\top \mathcal V(x). \) Hence, on such regions, UVP is gradient descent on \(g\) up to scaling: \(\mathcal P(x)=x-\frac{\eta_v}{2}\nabla g(x)\).

\begin{theorem}[Monotone decrease of violation]\label{thm:uvp_descent}
Assume \(\nabla g\) is \(L\)-Lipschitz on a set containing the iterates \(\{x_k\}\) and consider
\(x_{k+1}=x_k-\eta_v A^\top \mathcal V(x_k)\).
If \(\eta_v\in(0,4/L)\), then \(g(x_{k+1})\le g(x_k)\) for all \(k\) for which \(g\) is differentiable at \(x_k\),
in particular, for all \(k\) unless an iterate lands exactly on a constraint boundary. Moreover, \(\{g(x_k)\}\)
converges, and every accumulation point is stationary for \(g\).
\end{theorem}

\begin{proof}
Where \(g\) is differentiable, we write \(x_{k+1}=x_k-\alpha\nabla g(x_k)\) with \(\alpha=\eta_v/2\).
By the descent lemma for \(L\)-Lipschitz gradients,
\[
g(x_{k+1})\le g(x_k)-\alpha\Big(1-\tfrac{L\alpha}{2}\Big)\|\nabla g(x_k)\|_2^2.
\]
If \(\eta_v\in(0,4/L)\) then \(\alpha\in(0,2/L)\) and the coefficient is nonnegative, so \(g(x_{k+1})\le g(x_k)\).
Since \(g\ge 0\), \(\{g(x_k)\}\) converges. Summing the inequality over \(k\) shows
\(\sum_k \|\nabla g(x_k)\|_2^2<\infty\), hence \(\|\nabla g(x_k)\|\to 0\) along convergent subsequences. Any limit point satisfies \(\nabla g(\bar x)=0\), i.e.\ is stationary.
\end{proof}

\section{Instance Generator}\label{app:gen}
We generate cargo-demand instances \(q=\{q^{(i,j,k)}\}\) over \((i,j)\in \textit{TR}\) and \(k\in K\).

\paragraph{Training/test distribution (Gaussian).}
During training, we sample
\[
q^{(i,j,k)} \sim \mathcal N\!\big(\mu^{(i,j,k)},\,\sigma^{(i,j,k)}\big),
\qquad \sigma^{(i,j,k)}=\textit{CV}\cdot \mu^{(i,j,k)},
\]
so the coefficient of variation \(\textit{CV}=\sigma/\mu\) controls relative dispersion.

We draw the expected value (mean) as
\[
\mu^{(i,j,k)} \sim \mathcal U\!\big(0,\overline{\mu}^{(i,j,k)}\big),\qquad
\overline{\mu}=\frac{2\,\textit{UR}\cdot \mathbf 1^\top c}{\textit{NC}},
\]
where \(\textit{UR}\) sets total expected utilization (e.g., \(1.2\) corresponds to \(120\%\) of capacity) and
\(\textit{NC}\) spreads demand across \(\textit{TR}\) proportional to remaining transports.

\paragraph{Generalization distribution (uniform).}
For generalization testing, we sample from a uniform distribution with the same mean and variance:
\[
q^{(i,j,k)} \sim \mathcal U\!\big(\textit{lb}^{(i,j,k)},\,\textit{ub}^{(i,j,k)}\big).
\]
Using \(\mathbb E[q]=(\textit{lb}+\textit{ub})/2\) and \(\mathrm{Var}(q)=(\textit{ub}-\textit{lb})^2/12\), we set
\[
\textit{lb}^{(i,j,k)}=\mu^{(i,j,k)}-\sqrt{3}\,\sigma^{(i,j,k)},\qquad
\textit{ub}^{(i,j,k)}=\mu^{(i,j,k)}+\sqrt{3}\,\sigma^{(i,j,k)}.
\]

\paragraph{Nonnegativity.}
Demand is nonnegative, hence we use truncated samples \([q^{(i,j,k)}]_+\). This can introduce a mild left skew
when \(\sigma\) is large. 

\section{Multi-stage Stochastic Programs} \label{app:SP}
\subsection{Multi-stage Scenario Tree}
A scenario tree is a directed tree represented as \(T_\textit{ST} = (V_\textit{ST}, E_\textit{ST})\), where \( V_\textit{ST} \) is the set of nodes, each corresponding to a decision or uncertainty realization at a given stage. \( E_\textit{ST} \subseteq V_\textit{ST} \times V_\textit{ST} \) is the set of directed edges representing transitions between nodes over time.  The tree consists of:  
\begin{enumerate}
    \item A root node \( v_1 \in V_\textit{ST} \), representing the initial state at the first port. 
    \item Stages \( p = 1, 2, \dots, N_{P}-1 \), where each node \( v \) belongs to a stage \( p(v) \). We denote stages by $p$, as stages are equivalent to ports in a voyage.   
    \item Branching structure, where each node has child nodes that correspond to possible future realizations.
    \item A probability measure \(  P_\textit{ST}: V_\textit{ST} \to [0,1] \) assigning probabilities to nodes;     \[
   \sum_{v' \in \text{child}(v)} \mathbb{P}(v') =  \mathbb{P}(v), \quad \forall v \in V_\textit{ST}.
   \]  
   \item Scenario paths $\phi \in \mathcal{Z}$, which are root-to-leaf paths representing possible realizations of uncertainty over time.  
\end{enumerate}

Intuitively, the scenario tree discretely represents how uncertainty unfolds over time. Each branching point corresponds to a stochastic event, capturing possible realizations of uncertain parameters across the planning horizon. As we progress through stages, new uncertainty is realized, and decisions are made at each node based on available information. These decisions do not anticipate future realizations, but are optimized over the expected outcomes of future scenarios in the tree. In a perfect information formulation, this non-anticipativity is removed. Scenarios are thus no longer linked by shared decisions, and paths can be solved independently as if future uncertainty were known in advance. Averaging these objective values yields the expected value with perfect information.

\subsection{Mixed Integer Program}
We define the MPP under demand uncertainty as a multi-stage stochastic mixed integer program (MIP). We refer to Section \ref{sec:domain} and \refSuppMDP{} for sets and parameters introduced earlier. We include the following decision variables:
\begin{itemize}
    \item Vessel utilization: $\Tilde{u}^{b,d,\phi}_{\textit{tr},k} \in \mathbb{Z}_{\geq0}$
    \item Hatch overstowage: $\Tilde{\textit{ho}}^{\phi}_{p,b} \in \mathbb{Z}_{\geq0}$
    \item Makespan of cranes: $\Tilde{\textit{cm}}^{\phi}_{p} \in \mathbb{Z}_{\geq0}$
    \item Hatch movement: $\Tilde{\textit{hm}}^{\phi}_{p,b} \in \{0,1\}$
\end{itemize}

All integer constraints are relaxed linearly in the implementation. We also use a sufficiently large constant, denoted by big $M$, to impose logical constraints. 

The objective function \eqref{for:MIP_obj} maximizes the revenue with parameter \(\textit{rev}^{(i,j,k)} \in \mathbb{R}_{>0}\) and minimizes hatch-overstowage with parameter \( \textit{ct}^\textit{ho} \in \mathbb{R}_{>0} \) and crane moves costs with parameter \( \textit{ct}^\textit{cm} \in \mathbb{R}_{>0} \) over scenario paths $\phi \in \mathcal{Z}$ each with probability $\mathbb{P}_\phi$. We assume each scenario path has equal probability.

Constraint \eqref{for:MIP_demand} enforces that the onboard utilization cannot exceed the cargo demand $q$, whereas Constraint \eqref{for:MIP_capacity} limits each vessel location to the TEU capacity $c$ for each bay $b \in B$ and deck $d \in D$. In Constraint \eqref{for:hatch}, we indicate that hatches need to be opened if below deck cargo needs to be loaded or discharged. Based on these movements, Constraint \eqref{for:hatch_restow} models the amount of hatch overstowage in containers. Subsequently, we compute the target of crane moves $\overline{z}$ in Constraint \eqref{for:z_upper}, after which Constraint \eqref{for:long_crane} computes the excess number of crane moves $\Tilde{\textit{cm}}$. Additionally, we model the longitudinal and vertical stability in Constraints \eqref{for:lm} until \eqref{for:VS1}. First, we compute the longitudinal moment, vertical moment and total weight in Constraints \eqref{for:lm}, \eqref{for:vm} and \eqref{for:TW}, respectively. Second, Constraint \eqref{for:LS1} bounds \textit{lcg} between $\underline{\textit{lcg}}$ and $\overline{\textit{lcg}}$. Third,  Constraint \eqref{for:VS1} bounds \textit{vcg} between $\underline{\textit{vcg}}$ and $\overline{\textit{vcg}}$. Both \textit{lcg} and \textit{vcg} are linearized equivalents of the original Constraints \eqref{for:lcg} and \eqref{for:vcg}, respectively. Furthermore, we include non-anticipation in Constraint \eqref{for:nonanti} to prevent leveraging future demand realizations.

\begin{align}
\text{max } & 
\sum_{\phi \in \mathcal{Z}} \mathbb{P}_\phi \sum_{p\in {P}} \sum_{b \in{B}}\sum_{d\in {D}} \sum_{k\in {K}}\sum_{\textit{tr} \in \textit{TR}^\textit{+}(p)} \textit{rev}^{(i,j,k)} \Tilde{u}^{b,d,\phi}_{\textit{tr},k} \!-\! \textit{ct}^\textit{ho} \Tilde{\textit{ho}}^{\phi}_{p,b} \!-\! \textit{ct}^\textit{cm} \Tilde{\textit{cm}}^{\phi}_{p}  \label{for:MIP_obj}
\end{align}
\begin{align}
    \text{s.t. } & 
    \sum_{b \in {B}} \sum_{d \in {D}} \Tilde{u}^{b,d,\phi}_{\textit{tr},k} \leq q^{\phi}_{\textit{tr},k} \nonumber\\
    & \qquad\qquad\qquad \forall p \in {P},\; \textit{tr} \in \textit{TR}^\textit{OB}(p),\; k \in {K},\; \phi \in \mathcal{Z} \label{for:MIP_demand} \\
    & \sum_{k \in {K}} \sum_{\textit{tr} \in \textit{TR}^\textit{OB}(p)} \textit{teu}_{\textit{tr},k} \Tilde{u}^{b,d,\phi}_{\textit{tr},k} \leq c_{b,d} \nonumber\\
    & \qquad\qquad\qquad \forall p \in {P},\; b \in{B},\; d \in {D},\; \phi \in \mathcal{Z} \label{for:MIP_capacity} \\
    & \sum_{k \in {K}} \sum_{\textit{tr} \in \textit{TR}^{M}(p)} \Tilde{u}^{b,d_h,\phi}_{\textit{tr},k} \leq M \Tilde{\textit{hm}}^{\phi}_{p,b} \nonumber\\
    & \qquad\qquad\qquad \forall p \in {P},\; b \in{B},\; \phi \in \mathcal{Z} \label{for:hatch} \\
    & \sum_{k \in {K}} \sum_{\textit{tr} \in \textit{TR}^\textit{ROB}(p)} \Tilde{u}^{b,d_o,\phi}_{\textit{tr},k} - M(1 - \Tilde{\textit{hm}}^{\phi}_{p,b}) \leq \Tilde{\textit{ho}}^{\phi}_{p,b} \nonumber\\
    & \qquad\qquad\qquad \forall p \in {P},\; b \in{B},\; \phi \in \mathcal{Z} \label{for:hatch_restow} \\
    & {\overline{z}}^{\phi}_{p} = (1 + \delta^\textit{cm}) \frac{2}{|{B}|} \sum_{\textit{tr} \in \textit{TR}^M(p)} \sum_{k \in {K}} q^{\phi}_{\textit{tr},k} \nonumber\\
    & \qquad\qquad\qquad \forall p \in {P},\; \phi \in \mathcal{Z} \label{for:z_upper} \\
    & \sum_{b \in{b'}} \sum_{d \in {D}} \sum_{k \in {K}} \sum_{\textit{tr} \in \textit{TR}^{M}(p)} \Tilde{u}^{b,d,\phi}_{\textit{tr},k} - {\overline{z}}^{\phi}_{p} \leq \Tilde{\textit{cm}}^{\phi}_{p} \nonumber\\
    & \qquad\qquad\qquad \forall p \in {P},\; {b'} \in {B}',\; \phi \in \mathcal{Z} \label{for:long_crane} \\
    & \textit{tw}^{\phi}_{p} = \sum_{k \in {K}} w_k \sum_{\textit{tr} \in \textit{TR}^\textit{OB}(p)} \sum_{d \in {D}} \sum_{b \in{B}} \Tilde{u}^{b,d,\phi}_{\textit{tr},k} \nonumber\\
    & \qquad\qquad\qquad \forall p \in {P},\; \phi \in \mathcal{Z} \label{for:TW} \\
    & \textit{lm}^{\phi}_{p} = \sum_{b \in{B}} \textit{ld}_b \sum_{k \in {K}} w_k \sum_{\textit{tr} \in \textit{TR}^\textit{OB}(p)} \sum_{d \in {D}} \Tilde{u}^{b,d,\phi}_{\textit{tr},k} \nonumber\\
    & \qquad\qquad\qquad \forall p \in {P},\; \phi \in \mathcal{Z} \label{for:lm} \\
    & \textit{vm}^{\phi}_{p} = \sum_{d \in {D}} \textit{vd}_d \sum_{k \in {K}} w_k \sum_{\textit{tr} \in \textit{TR}^\textit{OB}(p)} \sum_{b \in{B}} \Tilde{u}^{b,d,\phi}_{\textit{tr},k} \nonumber\\
    & \qquad\qquad\qquad \forall p \in {P},\; \phi \in \mathcal{Z} \label{for:vm} \\
    & \underline{\textit{lcg}} \textit{tw}^{\phi}_{p} \leq \textit{lm}^{\phi}_{p} \leq \overline{\textit{lcg}} \textit{tw}^{\phi}_{p} \nonumber\\
    & \qquad\qquad\qquad \forall p \in {P},\; \phi \in \mathcal{Z} \label{for:LS1} \\
    & \underline{\textit{vcg}} \textit{tw}^{\phi}_{p} \leq \textit{vm}^{\phi}_{p} \leq \overline{\textit{vcg}} \textit{tw}^{\phi}_{p} \nonumber\\
    & \qquad\qquad\qquad \forall p \in {P},\; \phi \in \mathcal{Z} \label{for:VS1} \\
    & \Tilde{u}^{b,d,\phi'}_{\textit{tr},k} = \Tilde{u}^{b,d,\phi}_{\textit{tr},k} \nonumber\\
    & \qquad\qquad\qquad \forall p \in {P}, \textit{tr} \in \textit{TR}^{+}(p), k \in {K},\nonumber\\
    & \qquad\qquad\qquad b \in{B}, d \in {D}, \phi, \phi' \in \mathcal{Z} \mid q^{\phi}_{[p]} = q^{\phi'}_{[p]} \label{for:nonanti}
\end{align}

\section{DRL Implementation Details} \label{app:DRL}
\subsection{Loss Augmented RL}
We enforce \(A(s_t)x\le b(s_t)\) by augmenting the actor loss with a hinge penalty:
\[
\mathcal{L}(\theta)= \mathcal{L}_{\text{actor}}(\theta)+\mathcal{L}_{\text{feas}}(\theta),\qquad
\mathcal{L}_{\text{feas}}(\theta)=\mathbb{E}_t\!\left[\lambda_f^\top\big(A(s_t)x_\theta(s_t)-b(s_t)\big)_+\right],
\]
where \(\lambda_f\in\mathbb{R}^m_{>0}\) trades off return and feasibility. We consider these methods:
\begin{itemize}
\item \textbf{Penalty} (Pen-PPO, Pen-SAC): fixed \(\lambda_f\) tuned offline.
\item \textbf{Primal--dual} (Lag-PPO, Lag-SAC): learn \(\lambda_f\) as dual variables via gradient ascent on \(\mathcal{L}_{\text{feas}}\), implemented as a critic dual head \cite{dingNaturalPolicyGradient2020}.
\end{itemize}

\subsection{Feasibility Projection Optimization}
Each projection layer maps a raw action $x_{\mathrm{raw}}\in\mathbb{R}^n$ to an output $x\in\mathbb{R}^n$ using constraints $A\in\mathbb{R}^{m\times n}$, $b\in\mathbb{R}^m$, and non-negativity $x \geq 0$. We also split constraints into hard and soft row index sets $H,S$ to allow for violations.

\textbf{Unbiased Violation Projection.}
UVP performs $K$ projected gradient steps to reduce squared constraint violation:
\begin{equation}
\min_{x}\;\; \frac{1}{2}\big\|\big(Ax-b\big)_+\big\|_2^2
\quad\text{s.t.}\quad x\ge 0,
\end{equation}
implemented by iterations $x^{k+1}\leftarrow \Pi_{\{x\ge 0\}}\left(x^k-\eta A^\top(Ax^k-b)_+\right)$.

\textbf{CP Projection.}
The CP baseline projects onto hard constraints while allowing nonnegative slack $\xi$ on soft constraints:
\begin{equation}
\begin{aligned}
\min_{x,\xi}\quad & \frac{1}{2}\|x-x_{\mathrm{raw}}\|_2^2 + \lambda\|\xi\|_2^2\\
\text{s.t.}\quad & A_Hx \le b_H,\qquad A_Sx \le b_S + \xi,\qquad \xi\ge 0,\qquad x\ge 0,
\end{aligned}
\end{equation}

\textbf{$\alpha$-map with Chebyshev anchor.}
We first compute an interior anchor (Chebyshev-style) with the same soft-slack semantics:
\begin{equation}
\begin{aligned}
\max_{x,r,\xi}\quad & \beta r - \lambda\|\xi\|_2^2\\
\text{s.t.}\quad & r\ge 0,\; \xi\ge 0,\;
A_Hx + r d_H \le b_H,\;
A_Sx + r d_S \le b_S + \xi,\;
x\ge 0,
\end{aligned}
\end{equation}

where $d_i=\|A_{i:}\|_2$. Then an $\alpha$-map moves from the anchor $x_0$ toward $x_{\mathrm{raw}}$ while satisfying the relaxed RHS $b_{\mathrm{eff}}$ with $(b_{\mathrm{eff}})_S = b_S + \xi$.

\textbf{Frank--Wolfe Improvement (FW).}
FW computes an anchor $x_{\mathrm{feas}}$ via the CP projection above, then solves a linear minimization oracle for a critic gradient $g$. Then, it returns the convex combination $x=(1-\alpha)x_{\mathrm{feas}}+\alpha x'$.
\begin{equation}
\begin{aligned}
\min_{x',\xi}\quad & -g^\top x'\\
\text{s.t.}\quad & A_H x' \le b_H, \qquad x'\ge 0.
\end{aligned}
\end{equation}

\subsection{PPO Algorithm} 
PPO is an on-policy RL algorithm that seeks to maximize expected cumulative reward while enforcing stable policy updates via clipped importance sampling \cite{schulmanProximalPolicyOptimization2017}, as outlined in Algorithm \ref{alg:ppo}. The agent collects trajectories, computing \(n_\text{ppo}\)-step return to evaluate performance with \( V_{\theta}(s) \) as estimated state value:
\begin{equation}
G_t^{(n_\text{ppo})} = \sum_{k_\text{ppo}=0}^{n_\text{ppo}-1} \gamma^k r_{t+k_\text{ppo}} + \gamma^{n_\text{ppo}} V_{\theta}(s_{t+n_\text{ppo}}),  
\label{for:n_return}
\end{equation}

To reduce variance, we adopt Generalized Advantage Estimation (GAE): 
\begin{align}
\hat{A}_t^{\text{GAE}} &= \sum_{l_\text{ppo}=0}^{\infty} (\gamma \lambda)^{l_\text{ppo}} \delta_{t+l_\text{ppo}}, \label{for:gae1} \\ 
\delta_t &= r_t + \gamma V_{\theta}(s_{t+1}) - V_{\theta}(s_t). \label{for:gae2}
\end{align}
Here, \( \delta_t \) is the temporal difference (TD) residual, which quantifies the advantage of taking action \( x_t \) at state \( s_t \).

The actor is updated using the PPO clipped surrogate loss:
\begin{align}
\mathcal{L}_{\text{actor}}(\theta) &= \mathbb{E}_t \Big[ \min \big( \text{ratio}_t(\theta) \hat{A}_t^{\text{GAE}}, \text{clip}(\text{ratio}_t(\theta), 1 - \epsilon, 1 + \epsilon) \hat{A}_t^\text{GAE} \big) \Big], \label{for:ppo_actor}
\end{align}
where the probability ratio is defined as:
\begin{equation}
\text{ratio}_t(\theta) = \frac{\pi_{\theta}(x_t | s_t)}{\pi_{\theta_{\text{old}}}(x_t | s_t)}.
\end{equation}

The critic aims to minimize the squared TD error:
\begin{equation}
\mathcal{L}_{\text{critic}}(\theta) = \mathbb{E}_t \Big[ \big( V_{\theta}(s_t) - G_t^{(n_\text{ppo})} \big)^2 \Big]. 
\label{for:ppo_critic}
\end{equation}

Finally, the total PPO objective, including feasibility regularization from Equation \eqref{for:feas_loss}, is given by:
\begin{align}
\mathcal{L}(\theta) &= -\mathcal{L}_{\text{actor}}(\theta) + \lambda_c \mathcal{L}_{\text{critic}}(\theta) - \lambda_e \mathbb{E}_t\big[ \text{entropy}(\pi_{\theta})\big] +  \mathcal{L}_\text{feas}(\theta), 
\label{for:ppo_total_loss}
\end{align}
where \( \lambda_c \) and \( \lambda_e \) are weighting coefficients for the critic loss and entropy regularization, respectively.

\begin{algorithm}[h!]
\scriptsize
\caption{Proximal Policy Optimization (PPO)}
\label{alg:ppo}
\begin{algorithmic}[1]
\REQUIRE Model parameters $\theta$, steps $n$, learning rate $\eta$
\FOR{each gradient update}
    \FOR{each step $t$}
        \STATE Collect $n$-step trajectories $\{(s_t, x_t, r_t, s_{t+1})\}$
        \STATE Compute $n$-step returns $G_t^{(n_\text{ppo})}$
        \STATE Compute advantage estimates $\hat{A}_t^{\text{GAE}}$
    \ENDFOR
    \STATE Update parameters: $\theta \gets \theta + \eta \nabla_{\theta} \mathcal{L}(\theta)$
\ENDFOR
\RETURN Policy $\pi_\theta$
\end{algorithmic}
\end{algorithm}

\subsection{SAC Algorithm}  
Soft Actor-Critic (SAC) is an off-policy RL algorithm that optimizes both reward maximization and entropy to encourage efficient exploration \cite{haarnojaSoftActorCriticOffPolicy2018}, as outlined in Algorithm \ref{alg:sac}. It is based on maximum entropy RL, which aims to learn a stochastic policy that not only maximizes cumulative rewards but also maintains high entropy for robustness and stability. SAC leverages a soft Q-learning approach, using two Q-functions to mitigate overestimation bias, an entropy-regularized policy update, and an automatically adjusted temperature parameter to balance exploration and exploitation.

The algorithm maintains an actor network for policy learning, two Q-function critics for value estimation, a target Q-network for stable learning, and an adaptive temperature parameter to regulate entropy. The loss functions for standard SAC are derived from the Bellman backup equation and the policy gradient formulation, ensuring convergence to an optimal stochastic policy. We also include feasibility regularization from Equation \eqref{for:feas_loss} in the actor loss. 
\begin{itemize}
 \item Compute target Q-value:
\end{itemize}
\begin{align*}
Q_\text{target}(s_t,x_t) &= r_t + \gamma \mathbb{E}_{s_{t+1}, x_{t+1} \sim \pi} \Big[ \min_{l=1,2} Q_{\theta}^l(s_{t+1}, x_{t+1}) - \alpha \log \pi_{\theta}(x_{t+1} | s_{t+1}) \Big]
\end{align*}
\begin{itemize}
    \item Critic loss:
    \[
    \mathcal{L}_\text{critic}(\theta) = \mathbb{E} \Big[ (Q_{\theta}(s_t, x_t) - Q_\text{target}(s_t,x_t))^2 \Big]
    \]
    \item Actor loss:
    \[
    \mathcal{L}_\text{actor}(\theta) = \mathbb{E} \Big[ \alpha \log \pi_\theta(x_t | s_t) - Q_{\theta}(s_t, x_t) + \mathcal{L}_\text{feas}(\theta)  \Big]
    \]
    \item Temperature loss:
    \[
    \mathcal{L}_\alpha(\theta) = \mathbb{E} \Big[ -\alpha (\log \pi_\theta(x_t | s_t) + \text{entropy}_{\text{target}}) \Big]
    \]

\end{itemize}

This formulation ensures stability and encourages exploration by adapting the trade-off between exploitation and exploration dynamically.

\begin{algorithm}[h!]
\scriptsize
\caption{Soft Actor-Critic (SAC)}
\label{alg:sac}
\begin{algorithmic}[1]
\REQUIRE Parameters: actor \(\theta_\text{actor}\), critics \(\theta_{\text{critic}}^1, \theta_{\text{critic}}^2\), targets \((\theta_{\text{target}}^1, \theta_{\text{target}}^2) = (\theta_{\text{critic}}^1, \theta_{\text{critic}}^2)\), temperature \(\alpha\), learning rate actor $\eta_a$, learning rate critic $\eta_c$, learning rate temperature $\eta_\alpha$, soft update parameter $\tau$, replay buffer \(\mathcal{D}\).
\FOR{each iteration}
\FOR{each environment step $t$} 
    \STATE Sample action \(x_t \sim \pi_\theta(x_t | s_t)\)
    \STATE Perform transition \(s_{t+1} \sim \mathcal{T}(s_{t+1}|s_t,x_t)\)
    \STATE Observe reward \(r_t = \mathcal{R}(s_t,x_t)\),
    \STATE Store \((s_t, x_t, r_t, s_{t+1})\) in \(\mathcal{D}\).
\ENDFOR
\FOR{each gradient step}
    \STATE Sample a minibatch \((s_t, x_t, r_t, s_{t+1})\) from \(\mathcal{D}\).
    \STATE Compute target Q-value: $Q_\text{target}(s_t,x_t)$
    \STATE Update parameters: \\
    \quad \(\theta_{\text{critic}}^l \gets \theta_{\text{critic}}^l - \eta_c \nabla_l \mathcal{L}_\text{critic}(\theta) \text{ for } l \in \{1,2\}\)\\
    \quad \(\theta_{\text{actor}} \gets \theta_{\text{actor}} - \eta_a \nabla \mathcal{L}_\text{actor}(\theta)\) \\
    \quad \(\alpha \gets \alpha - \eta_\alpha \nabla \mathcal{L}_\alpha(\theta)\) \\
    \quad \(\theta_{\text{target}}^l \leftarrow \tau \theta_{\text{critic}}^l + (1 - \tau) \theta_{\text{target}}^l \text{ for } l \in \{1,2\}\)
\ENDFOR
\ENDFOR
\RETURN Policy $\pi_\theta$
\end{algorithmic}
\end{algorithm}

\subsection{Hyperparameters} 
The parameters of the MPP instances, presented in Table \ref{tab:env_params}, are chosen based on domain knowledge to generate problem instances reflecting real-world scenarios.

\begin{table}[hb!]
    \centering
    \scriptsize
    \caption{Instance parameters.} \label{tab:env_params}
    \begin{tabular}{lll}
        \toprule
        \textbf{Parameters} & \textbf{Symbol} & \textbf{Value} \\
        \midrule
        Voyage length & $N_P$ & 4 \\
        Number of bays & $N_B$ & 10 \\
        Cardinality deck set & $|D|$ & 2 \\
        Cardinality cargo set & $|K|$ & 12 \\
        Cardinality transport set & $|\textit{TR}|$ & 6 \\
        Vessel TEU & $\mathbf{1}^\top c$ & 1,000 \\
        Long term contract reduction & $\textit{LR}$ & 0.3 \\
        Utilization rate demand & $\textit{UR}$ & 1.1 \\ 
        lcg bounds & $(\underline{\textit{lcg}},\overline{\textit{lcg}})$ & (0.85,1.05) \\
        vcg bounds & $(\underline{\textit{vcg}},\overline{\textit{vcg}})$ & (0.95,1.15) \\
        Crane moves allowance & $\delta^\textit{cm}$ & 0.25 \\
        Overstowage costs & $\textit{ct}^\textit{ho}$ & 0.33 \\
        Crane move costs & $\textit{ct}^\textit{cm}$ & 0.5 \\
        \bottomrule
    \end{tabular}
\end{table}

Table \ref{tab:ppo_sac_hyperparameters} lists the hyperparameters of projected and loss-augmented PPO and SAC. Hyperparameter tuning is performed using Bayesian optimization. Only a selected subset of hyperparameters is optimized, each bounded by specified minimum and maximum values in Table \ref{tab:ppo_sac_hyperparameters}. The tuning process uses the objective value as the primary selection criterion, prioritizing feasibility whenever possible.

\begin{table*}[h!]
    \centering
    \scriptsize
    \caption{Hyperparameters for types of PPO and SAC. In the primal-dual (Lag) setting, $\eta$ is replaced by primal and dual learning rates, $\eta_{\mathrm{p}}$ and $\eta_{\mathrm{d}}$.} \label{tab:ppo_sac_hyperparameters}
    \begin{tabular}{llcccccc}
        \toprule
        \multicolumn{2}{c}{\textbf{Settings}} & \multicolumn{2}{c}{\textbf{Projection-based}}  & \multicolumn{2}{c}{\textbf{Loss Augmented}} &        \multicolumn{2}{c}{\textbf{Ranges}} \\
        \cmidrule(lr){1-2} \cmidrule(lr){3-4} \cmidrule(lr){5-6}
        \cmidrule(lr){7-8}
        \textbf{Hyperparameters} & \textbf{Symbol} & \textbf{PPO} & \textbf{SAC} & \textbf{PPO} & \textbf{SAC} & \textbf{Min} & \textbf{Max} \\
        \midrule
        \textbf{Actor Network} & & Attention & Attention  & Attention  & Attention \\
        \textbf{Number of Heads} & & 8 & 8  & 8  & 8  & 8 & 16\\
        \textbf{Hidden Layer Size} & & 512 & 512 & 512  & 512 & 128 & 512\\
        \textbf{Encoder Layers} & & 3 & 3  & 3  & 3 & 1 & 3\\
        \textbf{Decoder Layers} & & 3 & 3  & 3  & 3 & 1 & 3\\
        \textbf{Critic Network} & & $1\times \text{MLP}$ & $2\times \text{MLP}$ & $1\times \text{MLP}$ & $2\times \text{MLP}$ \\
        \textbf{Critic Layers} & & 4 & 4  & 4  & 4 & 2 & 4 \\
        \textbf{Target Network} & & No & Soft Update & No & Soft Update  \\
        \textbf{Target Update Rate} & $\tau$ & N/A & 0.005 & N/A & 0.005 \\
        \textbf{Dropout Rate} & & $0.009$ & $0.009$  &  $0.009$  &  $0.009$  & $0.001$ & $0.2$ \\
        \textbf{Max Policy Std.} & & $2.0$ & $2.0$  & $9.46$ & $9.46$ & $0.01$ & $10.0$ \\
        \midrule
        \textbf{Optimizer} & & Adam & Adam & Adam & Adam \\
        \textbf{Learning Rate (LR)} & $\eta$ & $1.47 \times 10^{-4}$ & $1.47 \times 10^{-4}$ & $1.65 \times 10^{-6}$ & $1.30 \times 10^{-5}$   & $1.0 \times 10^{-6}$ & $0.01$ \\
        \textbf{Primal LR (Lag)} & $\eta_\mathit{p}$ & - & - & $2.26 \times 10^{-6}$ & $1.03 \times 10^{-6}$  & $1.0 \times 10^{-6}$ & $0.01$\\
        \textbf{Dual LR (Lag)} & $\eta_\mathit{d}$ & - & - & $1.47  \times 10^{-5}$ & $7.57 \times 10^{-4}$ & $1.0 \times 10^{-6}$ & $0.01$ \\
        \textbf{Batch Size} & & 64 & 64 & 64 & 64 & 16 & 512 \\
        \textbf{Embedding Size} & & 128 & 128 & 128 & 128 & 64 & 256 \\
        \textbf{Discount Factor} & $\gamma$ & 0.99 & 0.99 & 0.99 & 0.99 \\
        \textbf{GAE} & $\lambda$ & 0.95 & N/A & 0.95 & N/A \\
        \textbf{Value Coefficient} & $\lambda_c$ & $0.50$  & N/A & $0.50$ & N/A \\
        \textbf{Entropy Coefficient} & $\lambda_e$ & $0.010$  & Learned & $0.010$ & Learned \\
        \textbf{Feasibility Penalty} & $\lambda_f$ & Vector  & Vector & Vector & Vector & $0.01$ & $10.0$ \\
        \textbf{Clip Parameter} & $\epsilon$ & 0.2 & N/A & 0.2 & N/A \\
        \textbf{Replay Buffer} & & No & Yes & No & Yes \\
        \textbf{Replay Buffer Size} & & N/A & $10^4$ & N/A & $10^4$ \\
        \textbf{Mini-batch Size} & & 32 & 32 & 32 & 32 & 16 & 64 \\
        \textbf{Update Epochs} & & 5 & 1 & 5 & 1 & 3 & 20 \\
        \textbf{Entropy Target} & & N/A & $-|{X}|$ & N/A & $-|{X}|$\\
        \midrule
        \textbf{Projection LR} & $\eta_v$ & $0.1$ & $0.1$ & N/A & N/A & $0.001$ & $0.1$\\
        \textbf{Projection Iterations} & & $100$ & $100$ & N/A & N/A & 50 & 500\\
        \textbf{Inference Threshold} & $\delta_v$ & $0.01$ & $0.01$ & N/A & N/A & $0.001$ & $0.1$ \\
        \midrule
        \textbf{Training Budget} & & $3.6 \times 10^{7}$ & $3.6 \times 10^{7}$ & $3.6 \times 10^{7}$ & $3.6 \times 10^{7}$\\
        \textbf{Validation Budget} & & $5.0 \times 10^{3}$ & $5.0 \times 10^{3}$ & $5.0 \times 10^{3}$ & $5.0 \times 10^{3}$\\
        \textbf{Validation Frequency} & & Every 20\% & Every 20\%  & Every 20\%  & Every 20\% \\
        \bottomrule
    \end{tabular}
\end{table*}

\end{document}

%% file: tables/results_table.tex
\begin{table}[t!]
\centering
\scriptsize
\setlength{\tabcolsep}{2pt}
\caption{Our AM-P vs. baselines, with \textbf{best trade-off}, methods significantly worse than UVP+R are marked with $^\dagger$ (paired two-sided Wilcoxon signed-rank test, Holm-adjusted $p<0.001$), and expected upper bound is marked with $^\ast$.}
\label{tab:comparison}
\begin{tabular}{lll rrrrr}
\toprule
\multicolumn{3}{c}{\textbf{Methods}} & \multicolumn{1}{c}{\textbf{Train}} & \multicolumn{2}{c}{\textbf{Testing ($\boldsymbol{N=30}$)}} & \multicolumn{2}{c}{\textbf{Generalization ($\boldsymbol{N=30}$)}} \\
\cmidrule(lr){1-3} \cmidrule(lr){4-4} \cmidrule(lr){5-6} \cmidrule(lr){7-8}
\multicolumn{1}{c}{\textbf{Model}} & \multicolumn{1}{c}{\textbf{Alg.}} & \multicolumn{1}{c}{\textbf{Proj.}} & \multicolumn{1}{r}{\textbf{Time}} & \multicolumn{1}{r}{\textbf{Obj. (\$)}} & \multicolumn{1}{r}{\textbf{Time}} & \multicolumn{1}{r}{\textbf{Obj. (\$)}} & \multicolumn{1}{r}{\textbf{Time}} \\
\midrule
\textbf{AM-P} & \textbf{SAC} & \textbf{UVP+R} & 1,857 & 1,494.4$\pm$90.7 & 4.8$\pm$0.3 & 1,509.5$\pm$108.8 & 4.8$\pm$0.3 \\
AM-P & SAC & CP & 70,181 & 1,493.9$\pm$89.9 & 3.1$\pm$0.2 & 1,508.9$\pm$109.5 & 3.3$\pm$0.3 \\
AM-P & SAC & FW & 83,948 & 1,488.8$\pm$92.5 & 11.5$\pm$0.4 & 1,509.7$\pm$106.8 & 11.5$\pm$0.3 \\
AM-P & SAC & $\alpha$ & 40,883 & 1,476.9$\pm$88.0 & 1.3$\pm$0.1 & 1,494.2$\pm$107.8 & 1.3$\pm$0.0 \\
AM$^\dagger$ & Pen-SAC & R & 1,528 & 1,307.1$\pm$96.3 & 1.8$\pm$0.3 & 1,323.9$\pm$80.6 & 1.8$\pm$0.3 \\
AM$^\dagger$ & Lag-SAC & R & 865 & 1,000.2$\pm$80.2 & 1.3$\pm$0.0 & 995.6$\pm$65.6 & 1.3$\pm$0.0 \\
\midrule
\textbf{AM-P} & \textbf{PPO} & \textbf{UVP+R} & 1,346 & 1,490.7$\pm$91.4 & 5.4$\pm$0.5 & 1,509.5$\pm$107.5 & 5.9$\pm$0.7 \\
AM-P & PPO & CP & 85,285 & 1,480.7$\pm$89.4 & 5.9$\pm$0.6 & 1,497.0$\pm$107.8 & 5.9$\pm$0.6 \\
AM-P & PPO & FW & 97,060 & 1,485.5$\pm$91.8 & 5.6$\pm$0.7 & 1,505.5$\pm$107.5 & 5.7$\pm$0.5 \\
AM-P & PPO & $\alpha$ & 74,583 & 1,431.7$\pm$96.3 & 1.3$\pm$0.1 & 1,461.4$\pm$92.4 & 1.3$\pm$0.1 \\
AM$^\dagger$ & Pen-PPO & R & 957 & 1,070.9$\pm$80.8 & 1.3$\pm$0.1 & 1,066.6$\pm$64.2 & 1.3$\pm$0.0 \\
AM$^\dagger$ & Lag-PPO & R & 1,041 & 979.5$\pm$74.5 & 1.3$\pm$0.0 & 975.2$\pm$58.8 & 1.3$\pm$0.0 \\
\midrule
SP$^\ast$ & CPLEX & -- & -- & 1,658.9$\pm$113.5 & 3,146.2$\pm$1,857.8 & 1,672.3$\pm$141.3 & 3,391.0$\pm$2,526.3 \\
\bottomrule
\end{tabular}
\end{table}

%% file: plots/ablations.tex
\newcommand{\PlotWithCI}[6]{%
  \addplot[name path=hi#1, draw=none]
    table[col sep=comma, x=iter, y expr=\thisrow{#5}+\thisrow{#6}] {#4};
  \addplot[name path=lo#1, draw=none]
    table[col sep=comma, x=iter, y expr=\thisrow{#5}-\thisrow{#6}] {#4};

  \addplot[forget plot, draw=none, fill=#2, fill opacity=0.18]
    fill between[of=hi#1 and lo#1];

  \addplot+[color=#2, #3, mark options={line width=0.7pt}]
    table[col sep=comma, x=iter, y=#5] {#4};
}

\newcommand{\PlotWithMinMax}[7]{%
  \addplot[name path=hi#1, draw=none, forget plot]
    table[col sep=comma, x=iter, y expr=\thisrow{#7}] {#4};
  \addplot[name path=lo#1, draw=none, forget plot]
    table[col sep=comma, x=iter, y expr=\thisrow{#6}] {#4};

  \addplot[forget plot, draw=none, fill=#2, fill opacity=0.18]
    fill between[of=hi#1 and lo#1];

  \addplot+[color=#2, #3]
    table[col sep=comma, x=iter, y=#5] {#4};
}

\newcommand{\PlotWithLogSE}[7]{%

  \addplot[name path=hi#1, draw=none, forget plot]
    table[col sep=comma, x=iter,
      y expr={
        ( (\thisrow{#5} > 0) && (\thisrow{#6} >= 0) ) ?
        ( (\thisrow{#5})*exp((#7)*(\thisrow{#6})/(\thisrow{#5})) ) :
        nan
      }] {#4};

  \addplot[name path=lo#1, draw=none, forget plot]
    table[col sep=comma, x=iter,
      y expr={
        ( (\thisrow{#5} > 0) && (\thisrow{#6} >= 0) ) ?
        ( (\thisrow{#5})*exp(-( #7)*(\thisrow{#6})/(\thisrow{#5})) ) :
        nan
      }] {#4};

  \addplot[draw=none, fill=#2, fill opacity=0.18, forget plot]
    fill between[of=hi#1 and lo#1];

  \addplot+[color=#2, #3,
    mark options={draw=#2, fill=#2, line width=0.7pt}]
    table[col sep=comma, x=iter, y=#5] {#4};
}

\begin{figure*}[t!]
\centering
\begin{tikzpicture}

\begin{groupplot}[
  group style={group size=2 by 2, horizontal sep=1.5cm, vertical sep=1.6cm},
]

\nextgroupplot[
  overview,
  title={(a) AM-P with SAC},
  xlabel={Violation norm (log, $\downarrow$)},
  ylabel={Profit (\$)},
  xmode=log,
  xmin=1e-3, xmax=6e5,
  ymin=-200, ymax=2000,
  ytick={0, 500, 1000,1500,2000},
  legend style={font=\tiny, draw=none, fill=none, at={(-0.3,1.75)}, anchor=north east},
  legend cell align=left,
]

\definecolor{cBlue}{HTML}{0072B2}
\definecolor{cSky}{HTML}{56B4E9}
\definecolor{cGreen}{HTML}{009E73}
\definecolor{cOrange}{HTML}{E69F00}
\definecolor{cVerm}{HTML}{D55E00}
\definecolor{cPurple}{HTML}{CC79A7}
\definecolor{cGray}{HTML}{4D4D4D}

\addplot+[
  ablationmark,
  color=cBlue,
  mark size=1.4,
  mark=*,
  mark options={draw=cBlue, fill=cBlue, opacity=0.65}
] table[col sep=comma, x=viol, y=obj] {data/uvp/UVP_R.csv};
\addlegendentry{UVP+R}

\addplot+[
  ablationmark,
  color=cSky,
  mark size=1.4,
  mark=*,
  mark options={draw=cSky, fill=cSky, opacity=0.65}
] table[col sep=comma, x=viol, y=obj] {data/uvp/UVP.csv};
\addlegendentry{UVP}

\addplot+[
  ablationmark,
  color=cOrange,
  mark size=1.4,
  mark=*,
  mark options={draw=cOrange, fill=cOrange, opacity=0.65}
] table[col sep=comma, x=viol, y=obj] {data/uvp/UVP_star.csv};
\addlegendentry{UVP*}

\addplot+[
  ablationmark,
  color=cGreen,
  mark size=1.4,
  mark=triangle*,
  opacity=0.65,
  mark options={draw=cGreen, fill=cGreen}
] table[col sep=comma, x=viol, y=obj] {data/uvp/pen.csv};
\addlegendentry{Pen}

\addplot+[
  ablationmark,
  color=cGreen,
  mark size=1.4,
  mark=triangle*,
  opacity=0.65,
  mark options={draw=cPurple, fill=cPurple}
] table[col sep=comma, x=viol, y=obj] {data/uvp/lag.csv};
\addlegendentry{Lag}

\addplot+[
  ablationmark,
  mark size=1.4,
  mark=square*,
  opacity=0.65,
] table[col sep=comma, x=viol, y=obj] {data/uvp/no_UVP.csv};
\addlegendentry{\st{UVP}}

\addplot+[
  ablationmark,
  mark size=1.4,
  mark=square*,
  opacity=0.65,
] table[col sep=comma, x=viol, y=obj] {data/uvp/no_CP.csv};
\addlegendentry{\st{CP} \;}

\addplot+[
  ablationmark,
  mark size=1.4,
  mark=square*,
  opacity=0.65,
] table[col sep=comma, x=viol, y=obj] {data/uvp/no_FW.csv};
\addlegendentry{\st{FW} \;}

\addplot+[
  ablationmark,
  mark size=1.4,
  mark=square*,
  opacity=0.65,
] table[col sep=comma, x=viol, y=obj] {data/uvp/no_alpha.csv};
\addlegendentry{\st{$\alpha$} \;}

\nextgroupplot[
  overview,
  title={(b) Training AM-P},
  xlabel={Train updates},
  ylabel={Profit (\textdollar)},
  xmin=0, xmax=800,
  ymin=1000, ymax=1800,
  ytick={1300,1500,...,2100},
  legend style={font=\tiny, draw=none, fill=none, at={(1.02,1)}, anchor=north west},
  axis y discontinuity=parallel,
]

\PlotWithMinMax{bJC}{darkblue}{solid, mark=none, thin, opacity=0.5}{data/train/sac_jc.csv}{obj}{min}{max}
\addlegendentry{SAC/JC}

\PlotWithMinMax{bNC}{darkred}{solid, mark=none, thin, opacity=0.5}{data/train/sac_nc.csv}{obj}{min}{max}
\addlegendentry{SAC/NC}

\PlotWithMinMax{bNC}{darkgreen}{solid, mark=none, thin, opacity=0.5}{data/train/ppo.csv}{obj}{min}{max}
\addlegendentry{PPO}

\nextgroupplot[
  overview,
  title={(c) Objective tuning UVP},
  xlabel={Iterations},
  ylabel={Profit (\textdollar)},
  ymin=600, ymax=1700,
  ytick={1000,1250, 1500,...,2000},
  xtick={0,50,...,300},
  legend to name=LegendCD,   
  legend style={draw=none, fill=none, font=\tiny, cells={anchor=west}},
  legend columns=4,
  axis y discontinuity=parallel,
]
\addlegendimage{solid, draw=darkyellow, mark=*, mark size=0.5, mark options={draw=darkyellow, fill=darkyellow}}
\addlegendentry{SAC/UVP/$\eta_v$}
\addlegendimage{dashed, draw=darkgreen, mark=square*, mark size=0.5, mark options={draw=darkgreen, fill=darkgreen}}
\addlegendentry{PPO/UVP/$\eta_v$}

\addlegendimage{dotted, draw=darkred, mark=*, mark size=0.5, mark options={draw=darkred, fill=darkred}}
\addlegendentry{SAC/UVP/$\tilde{\eta}_v = 0.1$}
\addlegendimage{dash dot, draw=darkblue, mark=square*, mark size=0.5, mark options={draw=darkblue, fill=darkblue}}
\addlegendentry{PPO/UVP/$\tilde{\eta}_v = 0.1$}

\PlotWithCI{cSVD}{darkyellow}{solid, mark=*, mark size=0.5,}{data/eta_obj/sac.csv}{obj}{err}
\PlotWithCI{cSVD}{darkgreen}{dashed, mark=square*, mark size=0.5,}{data/eta_obj/ppo.csv}{obj}{err}
\PlotWithCI{cSVD}{darkred}{dotted, mark=*, mark size=0.5,}{data/eta_obj/sac_static.csv}{obj}{err}
\PlotWithCI{cSVD}{darkblue}{dash dot, mark=square*, mark size=0.5,}{data/eta_obj/ppo_static.csv}{obj}{err}

\nextgroupplot[
  overview,
  title={(d) Feasibility tuning UVP},
  xlabel={Iterations},
  ylabel={Violation norm (log, $\downarrow$)},
  xtick={0,50,...,300},
  ymode=log,
  ymin=1e-5, ymax=3e4,
  ytick={1e-6,1e-4,1e-2,1e0,1e2,1e4},
  legend style={draw=none}, 
]
\PlotWithLogSE{SVDsac}{darkyellow}{solid, mark=*, mark size=0.5}{data/eta_viol/sac.csv}{viol}{err}{1}
\PlotWithLogSE{SVDppo}{darkgreen}{dashed, mark=square*, mark size=0.5}{data/eta_viol/ppo.csv}{viol}{err}{1}
\PlotWithLogSE{SVDsac}{darkred}{dotted, mark=*, mark size=0.5}{data/eta_viol/sac_static.csv}{viol}{err}{1}
\PlotWithLogSE{SVDppo}{darkblue}{dash dot, mark=square*, mark size=0.5}{data/eta_viol/ppo_static.csv}{viol}{err}{1}

\end{groupplot}

\node at ($(group c1r2.south west)!0.5!(group c2r2.south east) + (0,-1.2cm)$)
  {\pgfplotslegendfromname{LegendCD}};

\end{tikzpicture}
\caption{Ablation results on (a) projection, (b) training, (c,d) tuning UVP.}
\label{fig:ablation}
\end{figure*}

%% file: plots/sensitivity_analysis.tex
\begin{figure}[t!]
\centering
\begin{tikzpicture}

\begin{groupplot}[
  group style={
    group size=2 by 2,
    horizontal sep=1.6cm,
    vertical sep=1.6cm
  },
  width=0.48\textwidth,
  height=0.26\textwidth,
  ybar,
  enlarge x limits=0.18,
  title style={font=\scriptsize},
  tick label style={font=\scriptsize},
  label style={font=\scriptsize},
  legend style={legend columns=5, font=\scriptsize, draw=none, fill=none},
  grid=both,
  xmajorgrids=false,
  ymajorgrids=true,
  minor x tick num=0,
  minor y tick num=0,
  major grid style={line width=.4pt, draw=darkgray!40},
]

\nextgroupplot[
  title={(a) Objective (test, $N=30$)},
  ylabel={Profit (\textdollar)},
  xlabel={SP scenarios and DRL models},
  symbolic x coords={4,8,16,32,SAC,PPO},
  xtick={4,8,16,32,SAC,PPO},
  xticklabels={4,8,16,32,,},
  ymin=640, ymax=1850,
  ytick={1000,1250,...,1750},
  axis y discontinuity=parallel,
  legend to name=legendSens, 
  clip=false,
]

\addlegendimage{
  ybar,
  draw=black,
  pattern=dots,
  pattern color=darkgray,
  preaction={fill=darkred!60}
}
\addlegendentry{SP \;}

\addlegendimage{
  ybar,
  draw=black,
  pattern=north east lines,
  pattern color=darkgray,
  preaction={fill=darkyellow}
}
\addlegendentry{SAC/UVP+R \;}

\addlegendimage{
  ybar,
  draw=black,
  fill=darkgreen
}
\addlegendentry{PPO/UVP+R \;}

\addlegendimage{
  ybar,
  draw=black,
  pattern=vertical lines,
  pattern color=darkgray,
  preaction={fill=cyan!60}
}
\addlegendentry{Max Profit \;}

\addplot+[
  bar shift=0pt,
  bar width=5pt,
  pattern=dots,
  pattern color=darkgray,
  preaction={fill=darkred!60},
  draw=black,
  error bars/.cd,
     y dir=both, y explicit,
     error bar style={black},
] coordinates {
  (4,1666.2) +- (0,119.8)
  (8,1666.4) +- (0,115.0)
  (16,1664.7) +- (0,117.2)
  (32,1658.9) +- (0,113.5)
};

\addplot+[
  bar shift=0pt,
  bar width=5pt,
  fill=darkgreen,
  draw=black,
  error bars/.cd,
     y dir=both, y explicit,
     error bar style={black},
] coordinates {
  (PPO,1490.7) +- (0,91.4)
};

\addplot+[
  bar shift=0pt,
  bar width=5pt,
  pattern=north east lines,
  pattern color=darkgray,
  preaction={fill=darkyellow},
  draw=black,
  error bars/.cd,
     y dir=both, y explicit,
     error bar style={black},
] coordinates {
  (SAC,1494.4) +- (0,90.7)
};

\node[anchor=north, font=\scriptsize, xshift=8pt, yshift=19.5pt]
  at (axis cs:SAC,1) {DRL};

\nextgroupplot[
  title={(b) Amortized runtime (test, $N=30$)},
  ylabel={Time (log s)},
  xlabel={SP scenarios and DRL models},
  ymode=log,
  ymin=1, ymax=10000,
  ytick={1,10,100,1000,10000},
  yticklabels={$10^{0}$,$10^{1}$,$10^{2}$,$10^{3}$,$10^{4}$},
  symbolic x coords={4,8,16,32,SAC,PPO},
  xtick={4,8,16,32,SAC,PPO},
  xticklabels={4,8,16,32,,},
  scaled y ticks=false,
  clip=false,
]

\draw[black, thick, dashed]
  (rel axis cs:0,0.6945) -- (rel axis cs:1,0.6945);

\node[anchor=north east, font=\tiny]
  at (rel axis cs:0.98,0.6945) {10 min};

\addplot+[bar shift=0pt, bar width=5pt, pattern=dots, pattern color=darkgray, preaction={fill=darkred!60}, draw=black]
coordinates {(4,3.8) (8,18.8) (16,213.4) (32,3146.2)};

\addplot+[bar shift=0pt, bar width=5pt, fill=darkgreen, draw=black]
coordinates {(PPO,45.05)};

\addplot+[bar shift=0pt, bar width=5pt, pattern=north east lines, pattern color=darkgray, preaction={fill=darkyellow}, draw=black]
coordinates {(SAC,66.7)};

\node[anchor=north, font=\scriptsize, xshift=8pt, yshift=-4.5pt]
  at (axis cs:SAC,1) {DRL};

\nextgroupplot[
  title={(c) Demand spread (gen., $N=30$)},
  xlabel={Coefficient of Variation (CV)},
  ylabel={Profit (\$)},
  symbolic x coords={0.1,0.3,0.5,0.7,0.9},
  xtick=data,
  ymin=1300, ymax=1700,
  ytick={1400,1500,...,1800},
  axis y discontinuity=parallel,
]

\addplot+[
  bar shift=-3pt,
  bar width=5pt,
  pattern=north east lines,
  pattern color=darkgray,
  preaction={fill=darkyellow},
  draw=black,
  error bars/.cd,
    y dir=both, y explicit,
    error bar style={black},
] coordinates {
  (0.1,1513.7) +- (0,91.2)
  (0.3,1512.7) +- (0,95.9)
  (0.5,1509.5) +- (0,108.8)
  (0.7,1514.2) +- (0,126.3)
  (0.9, 1531.1) +- (0,148.3)
};

\addplot+[
  bar shift=+3pt,
  bar width=5pt,
  fill=darkgreen,
  draw=black,
  error bars/.cd,
    y dir=both, y explicit,
    error bar style={black},
] coordinates {
  (0.1,1515.0) +- (0,88.89)
  (0.3,1512.7) +- (0,95.7)
  (0.5,1509.5) +- (0,107.5)
  (0.7,1513.9) +- (0,123.3)
  (0.9,1526.1) +- (0,141.6)
};

\nextgroupplot[
  title={(d) Voyage length (gen., $N=30$)},
  xlabel={Number of Ports},
  ylabel={Profit (log \$)},
  symbolic x coords={4,6,8,10},
  xtick=data,
  ymode=log,
  ymin=500, ymax=10000,
  ytick={1000,2000,4000,8000},
  axis y discontinuity=parallel,
  scaled y ticks=false, log ticks with fixed point,
]

\addplot+[
  pattern=north east lines,
  pattern color=darkgray,
  preaction={fill=darkyellow},
  bar width=5pt,
  draw=black,
  error bars/.cd,
    y dir=both, y explicit,
    error bar style={black},
] coordinates {
  (4,1509.5) +- (0,108.8)
  (6,2326.0) +- (0,122.1)
  (8,3134.6) +- (0,140.4)
  (10,3963.9) +- (0,120.1)
};

\addplot+[
  fill=darkgreen,
  bar width=5pt,
  draw=black,
  error bars/.cd,
    y dir=both, y explicit,
    error bar style={black},
] coordinates {
  (4,1509.5) +- (0,107.5)
  (6,2324.10) +- (0,124)
  (8,3136.40) +- (0,140.5)
  (10,3967.27) +- (0,122)
};

\addplot+[
  pattern=vertical lines,
  pattern color=darkgray,
  preaction={fill=cyan!60},
  bar width=5pt,
  draw=black,
  error bars/.cd,
    y dir=both, y explicit,
    error bar style={black},
] coordinates {
  (4,1745.9) +- (0,187.9)
  (6,3071.1) +- (0,271.6)
  (8,4588.1) +- (0,328.7)
  (10,6315.6) +- (0,348.0)
};

\end{groupplot}

\node[anchor=north] at ($(current bounding box.south)$)
  {\pgfplotslegendfromname{legendSens}};

\end{tikzpicture}

\caption{Sensitivity analyses: (a) objective and (b) runtime per SP scenario and DRL model, (c) generalization across CV levels, and (d) across voyage lengths.}
\label{fig:comparison_subfigures}

\end{figure}